\documentclass[11pt]{article}

\usepackage[preprint]{acl}

\usepackage{times}
\usepackage{latexsym}
\usepackage{booktabs} %
\usepackage{array} %
\usepackage{multirow}
\usepackage{amsmath}
\usepackage{balance}

\usepackage[T1]{fontenc}

\usepackage[utf8]{inputenc}

\usepackage{microtype}

\usepackage{inconsolata}

\usepackage{graphicx}
\usepackage{xspace}
\usepackage{svg}
\usepackage{enumitem}
\setlist{itemsep=1pt, parsep=0pt, topsep=3pt, partopsep=0pt}

\usepackage{cleveref}
\usepackage{placeins}      %
\usepackage{fvextra}       %
\usepackage{algorithm}
\usepackage{algpseudocode}

\newcommand{\DCAblFullAbst}{49.8}
\newcommand{\DCAblFullAcc}{71.4}
\newcommand{\DCAblHybridAbst}{54.7}
\newcommand{\DCAblHybridAcc}{70.1}

\newcommand{\DCAblRerankAbst}{52.5}

\newcommand{\DCAblSig}{3}
\newcommand{\DCAblTests}{60}

\newcommand{\DCAbstainMax}{87.7}
\newcommand{\DCAbstainMin}{13.0}

\newcommand{\DCAccIter}{69.3}

\newcommand{\DCAccNoRag}{24.3}

\newcommand{\DCAccRag}{68.5}

\newcommand{\DCAttGeminiTwoFiveFlashAPIoff}{78.7}

\newcommand{\DCAttemptMax}{99.9}
\newcommand{\DCAttemptMin}{78.7}

\newcommand{\DCCautionIterLenient}{63.9}
\newcommand{\DCCautionIterStrict}{46.4}
\newcommand{\DCCautionNoRagLenient}{49.5}
\newcommand{\DCCautionNoRagStrict}{17.1}
\newcommand{\DCCautionRagLenient}{70.2}
\newcommand{\DCCautionRagStrict}{56.4}
\newcommand{\DCCrossLlamaAgree}{86.8}
\newcommand{\DCCrossLlamaKappa}{0.80}

\newcommand{\DCCrossQwenAgree}{92.8}
\newcommand{\DCCrossQwenKappa}{0.89}

\newcommand{\DCCrossQwenN}{1,008}
\newcommand{\DCDenyIter}{17.5}
\newcommand{\DCDenyNoRag}{32.4}
\newcommand{\DCDenyRag}{13.8}

\newcommand{\DCExhaustedShare}{1.9}

\newcommand{\DCFamCommandARagLenient}{72.0}
\newcommand{\DCFamCommandARagStrict}{69.7}

\newcommand{\DCFamDenyMoreNoRag}{9}
\newcommand{\DCFamDenyMoreRag}{2}

\newcommand{\DCFamNNoRag}{14}

\newcommand{\DCFamQwenThreeFiveRagLenient}{73.9}
\newcommand{\DCFamQwenThreeFiveRagStrict}{35.1}

\newcommand{\DCFleissKappa}{0.61}

\newcommand{\DCItAbstDelta}{-10.0}

\newcommand{\DCItDown}{30}

\newcommand{\DCItPairs}{40}

\newcommand{\DCJudgeSpread}{12.1}
\newcommand{\DCKAbstFive}{52.4}

\newcommand{\DCKAbstOneZero}{49.9}

\newcommand{\DCKAccFive}{70.4}

\newcommand{\DCKAccOneZero}{72.2}

\newcommand{\DCKModels}{8}

\newcommand{\DCLeakGPTFourOne}{15.5}
\newcommand{\DCLeakGPTOSS}{19.8}

\newcommand{\DCLooShiftGemmaFour}{-2.1}
\newcommand{\DCLooShiftGemmaThree}{-1.2}
\newcommand{\DCLooShiftQwenThree}{+3.5}
\newcommand{\DCLooShiftQwenThreeFive}{+3.6}
\newcommand{\DCLooShiftQwenThreeNext}{+3.0}

\newcommand{\DCMainItDown}{30}

\newcommand{\DCMainItUp}{0}
\newcommand{\DCMainRagDown}{0}
\newcommand{\DCMainRagPairs}{40}
\newcommand{\DCMainRagUp}{38}

\newcommand{\DCPremiseHitFive}{71.3}

\newcommand{\DCRagDown}{0}
\newcommand{\DCRagPairs}{40}
\newcommand{\DCRagUp}{38}

\newcommand{\DCRecallHitFive}{91.0}

\newcommand{\DCRespJudgeCapChars}{12{,}000}
\newcommand{\DCRespJudgeLongPct}{0.04}

\newcommand{\DCRhoLeaveFamilyOutRag}{0.84}

\newcommand{\DCRhoSelectiveAbstain}{-0.07}

\newcommand{\DCSelfPrefGemmaFourThreeOneBEightbit}{+1.0}
\newcommand{\DCSelfPrefGemmaFourThreeOneBEightbitHi}{+1.4}
\newcommand{\DCSelfPrefGemmaFourThreeOneBEightbitLo}{+0.5}
\newcommand{\DCSelfPrefQwenThreeEightTwoSevenBEightbit}{+1.8}
\newcommand{\DCSelfPrefQwenThreeEightTwoSevenBEightbitHi}{+2.2}
\newcommand{\DCSelfPrefQwenThreeEightTwoSevenBEightbitLo}{+1.4}
\newcommand{\DCSeverityGemmaFourThreeOneBEightbit}{+3.0}
\newcommand{\DCSeverityLagunaSTwoOneEightbit}{+4.9}
\newcommand{\DCSeverityQwenThreeEightTwoSevenBEightbit}{-7.9}

\newcommand{\DFAccMeanIter}{69.3}
\newcommand{\DFAccMeanNoRag}{24.3}
\newcommand{\DFAccMeanRag}{68.5}
\newcommand{\DFAttMeanIter}{97.9}

\newcommand{\DFDepthIter}{44.9}
\newcommand{\DFDepthIterLower}{11}
\newcommand{\DFDepthKFive}{56.8}
\newcommand{\DFDepthKTen}{51.7}
\newcommand{\DFDepthModels}{11}
\newcommand{\DFDepthNoRag}{12.2}
\newcommand{\DFItDown}{30}
\newcommand{\DFItPairs}{40}

\newcommand{\DFMeanIter}{46.4}
\newcommand{\DFMeanNoRag}{17.1}
\newcommand{\DFMeanRag}{56.4}

\newcommand{\DFNAccAbst}{40}

\newcommand{\DFNModels}{27}

\newcommand{\DFNReleaseFamily}{14}
\newcommand{\DFNReleaseModel}{27}

\newcommand{\DFNTotal}{20}

\newcommand{\DFPAccAbst}{0.08}
\newcommand{\DFPAccAbstPartial}{0.04}
\newcommand{\DFPAccAttAbst}{0.68}

\newcommand{\DFPAccTotal}{0.005}
\newcommand{\DFPActive}{0.37}

\newcommand{\DFPReleaseFamily}{0.45}
\newcommand{\DFPReleaseModel}{0.42}

\newcommand{\DFPTotal}{0.31}

\newcommand{\DFPipeBm}{59.4}
\newcommand{\DFPipeDense}{52.2}
\newcommand{\DFPipeFull}{49.8}
\newcommand{\DFPipeFullBelow}{12}
\newcommand{\DFPipeHybrid}{54.7}
\newcommand{\DFPipeModels}{12}
\newcommand{\DFPipeRerank}{52.5}
\newcommand{\DFPipeRewrite}{52.5}

\newcommand{\DFRagPairs}{40}
\newcommand{\DFRagUp}{38}

\newcommand{\DFRhoAbstRelease}{-0.03}
\newcommand{\DFRhoAccAbst}{-0.28}
\newcommand{\DFRhoAccAbstPartial}{-0.33}
\newcommand{\DFRhoAccAttAbst}{-0.07}
\newcommand{\DFRhoAccRelease}{+0.60}

\newcommand{\DFRhoAccTotal}{+0.60}
\newcommand{\DFRhoActive}{+0.21}

\newcommand{\DFRhoReleaseFamily}{-0.22}
\newcommand{\DFRhoReleaseModel}{-0.16}

\newcommand{\DFRhoTotal}{-0.24}

\newcommand{\DFThinkAccModels}{13}
\newcommand{\DFThinkAccNoRag}{+1.1}
\newcommand{\DFThinkAccRag}{+3.4}
\newcommand{\DFThinkModels}{13}
\newcommand{\DFThinkNoRagDown}{8}

\newcommand{\DFThinkNoRagMeanMag}{14}
\newcommand{\DFThinkNoRagMeanOther}{-10.5}
\newcommand{\DFThinkNoRagMeanQwenThree}{-18.0}

\newcommand{\DFThinkNoRagP}{0.001}

\newcommand{\DFThinkRagMean}{-0.9}

\newcommand{\DFThinkRagMeanOther}{-11.6}
\newcommand{\DFThinkRagMeanQwenThree}{+11.6}
\newcommand{\DFThinkRagNOther}{7}
\newcommand{\DFThinkRagNQwenThree}{6}
\newcommand{\DFThinkRagP}{0.68}

\newcommand{\DFTradeAttMin}{95}
\newcommand{\DFTradeBothUp}{33}
\newcommand{\DFTradeCarDownAttUp}{1}
\newcommand{\DFTradeCarUpAttDown}{6}
\newcommand{\DFTradeNoRagAtt}{90.6}
\newcommand{\DFTradeNoRagBestAtt}{95.3}
\newcommand{\DFTradeNoRagBestCar}{21.0}
\newcommand{\DFTradeNoRagBestModel}{Qwen3 32B}
\newcommand{\DFTradeNoRagBestSetting}{thinking disabled}
\newcommand{\DFTradeNoRagCar}{17.1}
\newcommand{\DFTradeRagAtt}{97.3}
\newcommand{\DFTradeRagBestAtt}{95.7}
\newcommand{\DFTradeRagBestCar}{83.3}
\newcommand{\DFTradeRagBestModel}{Gemma 4 31B}
\newcommand{\DFTradeRagBestSetting}{thinking disabled}
\newcommand{\DFTradeRagCar}{56.4}

\newcommand{\DFAccMeanIterStrict}{69.3}

\newcommand{\DFAccMeanRagStrict}{68.5}

\newcommand{\DFItDownStrict}{30}
\newcommand{\DFItPairsStrict}{40}

\newcommand{\DFMeanIterStrict}{46.4}

\newcommand{\DFNModelsStrict}{27}

\newcommand{\DFPipeModelsStrict}{12}

\newcommand{\DFRagPairsStrict}{40}
\newcommand{\DFRagUpStrict}{38}

\newcommand{\DFRhoAccTotalStrict}{+0.60}

\newcommand{\DFScenDefaultStrict}{50.9}
\newcommand{\DFScenModelsStrict}{10}
\newcommand{\DFScenQonlyHigherStrict}{10}
\newcommand{\DFScenQonlyStrict}{57.4}

\newcommand{\DFThinkModelsStrict}{13}
\newcommand{\DFThinkNoRagDownStrict}{8}

\newcommand{\DFThinkNoRagMeanMagStrict}{14}

\newcommand{\DFItDownLenient}{22}

\newcommand{\DFNReleaseModelLenient}{27}

\newcommand{\DFPReleaseModelLenient}{0.01}

\newcommand{\DFRagPairsLenient}{40}
\newcommand{\DFRagUpLenient}{32}

\newcommand{\DFRhoReleaseModelLenient}{+0.47}

\newcommand{\DFThinkNoRagDownLenient}{3}

\newcommand{\DFThinkNoRagMeanLenient}{+1.9}

\newcommand{\DFThinkNoRagUpLenient}{5}

\newcommand{\HAAnnotators}{3}
\newcommand{\HADeclineHumanKappa}{0.67}
\newcommand{\HADeclineHumanKappaHi}{0.80}
\newcommand{\HADeclineHumanKappaLo}{0.51}
\newcommand{\HADeclineJudgeKappa}{0.82}
\newcommand{\HADeclineJudgeKappaHi}{0.90}
\newcommand{\HADeclineJudgeKappaLo}{0.73}
\newcommand{\HAHybridMajDeny}{7}
\newcommand{\HAHybridN}{9}
\newcommand{\HAOneHumanKappa}{0.57}
\newcommand{\HAOneHumanKappaHi}{0.68}
\newcommand{\HAOneHumanKappaLo}{0.45}
\newcommand{\HAOneJudgeKappa}{0.74}
\newcommand{\HAOneJudgeKappaHi}{0.82}
\newcommand{\HAOneJudgeKappaLo}{0.66}
\newcommand{\HAOneJudgeKappaRest}{0.79}
\newcommand{\HAOneMajAgree}{96}

\newcommand{\HAOneMajKappa}{0.93}
\newcommand{\HAOneMajKappaHi}{1.00}
\newcommand{\HAOneMajKappaLo}{0.85}
\newcommand{\HAOneN}{70}
\newcommand{\HATwoHumanKappa}{0.42}
\newcommand{\HATwoHumanKappaHi}{0.62}
\newcommand{\HATwoHumanKappaLo}{0.21}
\newcommand{\HATwoHumanMean}{0.51}
\newcommand{\HATwoJudgeKappa}{0.66}
\newcommand{\HATwoJudgeKappaHi}{0.79}
\newcommand{\HATwoJudgeKappaLo}{0.50}
\newcommand{\HATwoN}{30}
\newcommand{\HATwoPanelMean}{0.48}

\newcommand{\DMKnownGemmaFour}{20.2}

\newcommand{\DMKnownQwenThreeTwoThreeFive}{17.3}

\newcommand{\DMLostQwenThreeTwoThreeFive}{194}

\newcommand{\DMPanelFacts}{2,391}
\newcommand{\DMPanelGLM}{7.4}
\newcommand{\DMPanelGemmaFour}{9.0}
\newcommand{\DMPanelQwenThreeFive}{9.0}

\newcommand{\DMPostChunksMax}{64}
\newcommand{\DMPostChunksMedian}{2}

\newcommand{\DMStyleFinal}{0.57}
\newcommand{\DMStyleFinalFull}{0.80}

\newcommand{\DMStyleQwenThreeFive}{0.52}
\newcommand{\DMStyleQwenThreeTwoThreeFive}{0.65}
\newcommand{\DMTarget}{100}
\newcommand{\DMVerGemmaDrafts}{213}
\newcommand{\DMVerGemmaHitFive}{93.3}
\newcommand{\DMVerGemmaRej}{11}
\newcommand{\DMVerGemmaRejPct}{5.2}
\newcommand{\DMVerLagunaDrafts}{144}
\newcommand{\DMVerLagunaHitFive}{80.0}

\newcommand{\DTNoRagAtoD}{372}

\newcommand{\DTNoRagDtoA}{51}
\newcommand{\DTNoRagLostToDeny}{54.6}
\newcommand{\DTNoRagModels}{13}

\newcommand{\DTNoRagNetMean}{+8.6}
\newcommand{\DTNoRagNetP}{0.001}
\newcommand{\DTNoRagPairs}{3,404}
\newcommand{\DTNoRagSigAtoD}{9}
\newcommand{\DTNoRagSigAtoX}{5}

\newcommand{\DTRagAtoD}{315}

\newcommand{\DTRagDtoA}{133}

\newcommand{\DTRagSigAtoD}{6}

\newcommand{\DTRagSigDtoA}{1}
\newcommand{\DTRagSigXtoA}{5}

\newcommand{\DAARoneRagNativeAcc}{44.9}

\newcommand{\DAARoneRagNativeAtt}{73.1}

\newcommand{\DAARoneRagNativeCar}{87.7}

\newcommand{\DAFrontierAtt}{78.7}
\newcommand{\DAFrontierCar}{87.7}
\newcommand{\DAFrontierModel}{Gemini 2.5 Flash}
\newcommand{\DAMainAccAll}{68.5}
\newcommand{\DAMainAttMin}{78.7}
\newcommand{\DAMainN}{40}

\newcommand{\DAQSevenRagAcc}{57.9}
\newcommand{\DAQSevenRagAccAtt}{60.5}

\newcommand{\DAQSevenRagCar}{57.0}

\newcommand{\DAQThreeRagAcc}{49.1}
\newcommand{\DAQThreeRagAccAtt}{50.3}

\newcommand{\DAQThreeRagCar}{35.7}

\newcommand{\DAQThreeRagNativeAtt}{75.9}
\newcommand{\DAQThreeRagNativeCar}{75.7}

\newcommand{\DATASevenDeclinedBaseAcc}{52.4}
\newcommand{\DATASevenDeclinedMainAcc}{64.5}
\newcommand{\DATASevenDeclinedN}{401}

\newcommand{\DATASevenRagAcc}{24.5}
\newcommand{\DATASevenRagAccAtt}{57.4}
\newcommand{\DATASevenRagAtt}{42.7}

\newcommand{\DATASevenRagCar}{89.7}

\newcommand{\DATAThreeDeclinedBaseAcc}{43.2}
\newcommand{\DATAThreeDeclinedMainAcc}{65.7}
\newcommand{\DATAThreeDeclinedN}{395}

\newcommand{\DATAThreeRagAcc}{22.7}
\newcommand{\DATAThreeRagAccAtt}{52.2}
\newcommand{\DATAThreeRagAtt}{43.6}

\newcommand{\DATAThreeRagCar}{89.3}
\newcommand{\DATAThreeRagCarHi}{92.3}
\newcommand{\DATAThreeRagCarLo}{85.3}

\newcommand{\DBConfigs}{40}
\newcommand{\DBDraws}{1,000}

\newcommand{\DBRhoLo}{0.97}
\newcommand{\DBRhoHi}{0.99}
\newcommand{\DBTopShare}{91}
\newcommand{\DBTopConfig}{Gemma 4 31B}

\newcommand{\DBRankWidthMed}{6}

\newcommand{\DEConfigs}{9}
\newcommand{\DEModels}{6}

\newcommand{\DENatRBMin}{93}
\newcommand{\DENatRBMax}{99}
\newcommand{\DEAnsRefDCMin}{5}
\newcommand{\DEAnsRefDCMax}{19}
\newcommand{\DEAnsRefRBMin}{33}
\newcommand{\DEAnsRefRBMax}{78}
\newcommand{\DENatUnans}{243}
\newcommand{\DENatAns}{299}
\newcommand{\DEGoldHit}{46}

\newcommand{\name}{\textsc{DeclineCraft}\xspace} %

\title{\name: A RAG Abstention Benchmark for Jointly Measuring Combined Retrieval and Parametric Knowledge}

\author{Atharva Thakur \quad Abulhair Saparov \quad Bruno Ribeiro \\
  Purdue University \\
  \texttt{\{thakur27, asaparov, ribeirob\}@purdue.edu}}

\begin{document}
\maketitle
\begin{abstract}
Reliable retrieval-augmented generation (RAG) requires recognizing when neither retrieved documents nor parametric knowledge supports an answer.
We introduce \name, a dynamic abstention benchmark whose unanswerable questions target \textit{Minecraft} feature additions after a corpus cut-off, with audits of possible parametric exposure.\footnote{Code and data are freely available at: \url{https://github.com/Thakur-Atharva/declinecraft-benchmark}}
Later cut-offs support evaluation of future models.
Across \DFNModelsStrict{} LLMs, we measure Correct Abstention Rate (CAR; how often models acknowledge insufficient knowledge on unanswerable questions), Answerable Attempt Rate and Answer Accuracy.
We find that (1) single-shot retrieval significantly raises CAR in \DFRagUpStrict{} of \DFRagPairsStrict{} configurations (model and thinking setting) versus no retrieval.
(2) Iterative RAG, in which the model self-assesses its context and can request up to two further retrieval rounds, significantly lowers CAR versus single-shot RAG in \DFItDownStrict{} of \DFItPairsStrict{} configurations (mean \DFMeanIterStrict\%), with little change in Answer Accuracy (\DFAccMeanRagStrict\% to \DFAccMeanIterStrict\%).
Adding query rewriting and reranking to hybrid (dense \& sparse) retrieval lowers CAR for all \DFPipeModelsStrict{} models tested.
(3) Larger models answer more accurately (Spearman $\rho = \DFRhoAccTotalStrict$ with total parameters), but neither model size nor release date shows a significant association with CAR.
(4) Without retrieval, enabling thinking on the same weights lowers CAR significantly for \DFThinkNoRagDownStrict{} of \DFThinkModelsStrict{} models (by \DFThinkNoRagMeanMagStrict{}\% points on average), and the responses shift from acknowledged uncertainty to premise denial.
\end{abstract}

\section{Introduction}

A player fleeing skeletons in \textit{Minecraft Java Edition} asks an assistant whether copper horse armor would let a horse survive three arrows.
Answering requires combining the rules for arrow damage, horse health points, and the protection supplied by this particular armor in a specific game edition.
Suppose copper horse armor was introduced after both the model's training and its retrieval corpus.
Passages about horses, skeletons and other armor types may be relevant, yet the key fact about the horse's armor is missing.
The assistant must acknowledge this knowledge gap.
For instance, when asked for the armor's protection, \texttt{Qwen3 32B} with thinking enabled instead declares: ``Since `copper horse armor' is not a recognized item in the game, it would provide 0 armor points. If you meant golden horse armor, it provides 7 armor points.'' (\Cref{tab:example}). This response treats a presupposition it cannot verify as false~\citep{kim-etal-2021-linguist,kim-etal-2023-qa}, a {\em premise denial}, then substitutes a different item for the one the user asked about.

For RAG systems, recognizing such a knowledge gap must remain part of the decision to answer as retrieval and thinking proceed.
This matters most when the answers are consumed by AI agents, which may not notice that a premise denial is unsupported by the model's available knowledge.
A human user, by contrast, may know that a new law was just passed, or that {\em copper horse armor} was added to \textit{Minecraft}, when posing their query.
Evaluation must therefore distinguish an admission of missing knowledge from a premise denial that the model cannot verify.

\begin{table}[t]
\centering
\small
\renewcommand{\arraystretch}{1.12}
\begin{tabular}{@{}>{\raggedright\arraybackslash}p{0.25\columnwidth}@{\hspace{0.03\columnwidth}}>{\raggedright\arraybackslash}p{0.72\columnwidth}@{}}
\toprule
\multicolumn{2}{@{}p{\columnwidth}@{}}{\textbf{Answerable.} \textit{``I am taming a horse in the plains by feeding it golden carrots until heart particles appear, then I equip it with a saddle and armor. If I encounter a skeleton horse, what item will it drop when killed?''}} \\[2pt]
\multicolumn{2}{@{}p{\columnwidth}@{}}{Reference answer: bones, documented in $C_{\mathrm{pre}}$.} \\
\midrule
\multicolumn{2}{@{}p{\columnwidth}@{}}{\textbf{Unanswerable.} \textit{``I ride my horse across the plains, holding the jump control to charge a high leap over the fence gate blocking my path. If I equip copper horse armor, how many armor points will it provide?''}} \\[2pt]
\multicolumn{2}{@{}p{\columnwidth}@{}}{The answer (4 points) is documented in $C_{\mathrm{post}}$; the retrieved $C_{\mathrm{pre}}$ pages list leather, iron, gold and diamond horse armor.} \\
\midrule
\textbf{Abstention} \newline \texttt{Gemma 4 31B}
  & ``The provided context does not contain the answer. I would need to know if copper horse armor exists in the game, as the provided text only lists leather, iron, gold, and diamond variants.'' \\[4.8em]
\textbf{Premise denial} \newline \texttt{Qwen3 32B} (thinking)
  & ``\ldots\ Since `copper horse armor' is \textbf{not a recognized item in the game}, it would provide \textbf{0 armor points}. \textbf{If you meant golden horse armor, it provides 7 armor points}.'' \\[4.8em]
\textbf{Confabulation} \newline \texttt{Llama 3.3 70B}
  & ``\ldots\ Copper horse armor provides \textbf{8 armor points} for your horse.'' \\
\bottomrule
\end{tabular}
\caption{Two benchmark questions and real responses to the unanswerable one; bold highlights unsupported claims. Copper horse armor was documented on the wiki only from July 2025, after the corpus cut-off. The first response acknowledges missing knowledge; the second is a premise denial followed by a substituted item; the third confabulates a numerical answer. Only the first is a correct abstention. The first two use single-shot RAG; the third is without retrieval (with RAG, \texttt{Llama 3.3 70B} abstains).}
\label{tab:example}
\end{table}

Distinguishing abstention from premise denial matters for retrieval-augmented generation (RAG) wherever documentation and the world it describes can change \citep{barnett2024sevenfailurepointsengineering,xu2024knowledgeconflictsllmssurvey}.
RAG systems can combine retrieved documents with \emph{parametric knowledge}, the information stored in a model's weights \citep{lewis2021retrievalaugmentedgenerationknowledgeintensivenlp}.
Unless explicitly asked about a historical version, a system should allow for questions about a time after both the corpus cut-off and the model's knowledge cut-off.
An item absent from an earlier version may exist when the user asks about it; absence from the available sources does not establish present-day nonexistence.
A reliable system should therefore \emph{abstain} when these sources together cannot support an answer, identifying the knowledge gap, without categorically rejecting a premise it cannot verify.

Evaluating this behavior is challenging: an answer missing from retrieved passages may still be available in the model's weights.
Context-grounded benchmarks instruct models to answer only from supplied documents \citep{song2025trustalign,ming2025faitheval,muhamed2025refusalbench}.
UAEval4RAG's out-of-database questions include facts from newer news absent from its database, under a database-grounded answering protocol \citep{peng-etal-2025-unanswerability}.
AbstentionBench evaluates, without retrieval, questions whose answers changed after the models' knowledge cut-offs \citep{kirichenko2025abstentionbench}.
Our study combines a time-partitioned corpus, whose later part postdates the models' knowledge cut-offs, with related but insufficient retrieved evidence, and scores premise denial separately from acknowledged uncertainty.

\begin{figure*}[t]
    \centering
    \includegraphics[width=\textwidth]{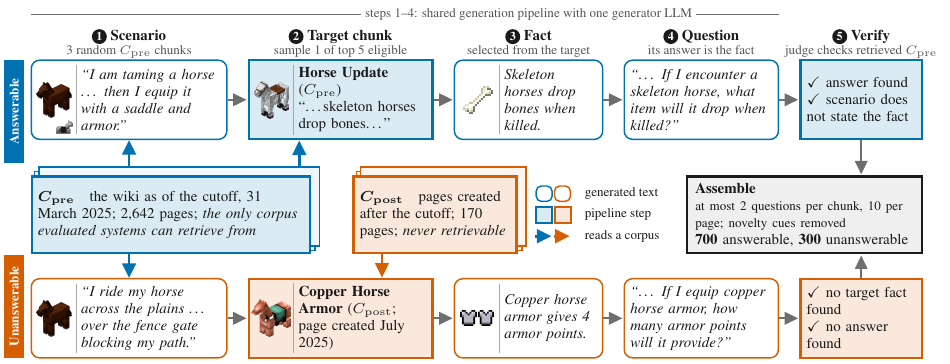}
    \caption{Building \name. A shared pipeline generates a player scenario from the pre-cut-off corpus $C_\text{pre}$, samples one of the five highest-ranked eligible target chunks, selects a fact and appends a question requiring the selected fact. The target comes from $C_\text{pre}$ for answerable questions and the post-cut-off corpus $C_\text{post}$ for unanswerable questions. A separate verifier checks whether retrieved $C_\text{pre}$ passages support the answer. Both examples are real benchmark items.}
    \label{fig:method}
\end{figure*}

To evaluate this setting, we introduce \name, a dynamic benchmark of 700 answerable and 300 unanswerable questions built from the update history of \textit{Minecraft} \citep{minecraftwiki2026}.
The interacting game mechanics of \textit{Minecraft} and its feature additions over time create facts that can only be learned from new wiki pages: e.g., knowing how existing armor works does not establish the protection of a newly introduced material against a pre-existing weapon.
We construct a retrieval corpus ($C_\text{pre}$) from wiki pages as they stood at a corpus cut-off and generate unanswerable questions from pages created later ($C_\text{post}$).
Answerable questions use the same generation pipeline as unanswerable ones, with their target facts drawn from the earlier corpus (\Cref{fig:method}).
We audit our unanswerable questions against the earlier documentation. To assess possible parametric exposure, we use published knowledge cut-offs and a \emph{leakage audit}. The leakage audit tests whether models answer the unanswerable questions correctly without retrieval.
The temporal design establishes a knowledge gap when the required facts postdate both parametric knowledge and the RAG corpus.
The benchmark is dynamic: it can be regenerated with a later corpus cut-off as new game content becomes available.

We evaluate \DFNModels{} LLMs from several families under three retrieval conditions, with thinking enabled and disabled where possible.
\emph{Single-shot RAG} retrieves five wiki passages (\emph{chunks}) once, combining keyword and embedding search.
\emph{Iterative RAG} lets the model request up to two further retrieval rounds when it judges the chunks insufficient (\Cref{sec:retrieval-results}).
We measure \emph{Correct Abstention Rate} (CAR) as the share of unanswerable questions on which a model acknowledges that it lacks the answer.
Our experiments yield three main findings:
\begin{itemize}[leftmargin=*]
    \item \textbf{Retrieval helps models recognize missing knowledge, but additions to the retrieval pipeline erode that benefit.}
    Single-shot RAG raises CAR for almost every configuration.
    Iterative RAG, ten instead of five chunks per query, and query rewriting with reranking each lower it again (\Cref{sec:retrieval-results}).
    \item \textbf{Newer and larger models do not reliably acknowledge knowledge gaps.}
    CAR has no significant association with release date or parameter count.
    Recent families such as Qwen3.5 and Qwen3-Next often deny the existence of the item or feature mentioned in the question (\Cref{sec:res-size}).
    \item \textbf{Thinking can replace uncertainty with premise denial.}
    Without retrieval, enabling thinking significantly lowers CAR for \DFThinkNoRagDownStrict{} of the \DFThinkModels{} models with switchable thinking and raises it for none.
    The lost abstentions most often become premise denials.
    With retrieval, the effect of thinking differs by model (\Cref{sec:res-thinking}).
\end{itemize}

\section{Related Work}
\paragraph{Abstention and selective QA.}
SQuAD 2.0 evaluates abstention when a passage lacks the answer \citep{rajpurkar-etal-2018-know}.
Selective QA studies when to withhold answers to control error among answered questions \citep{kamath-etal-2020-selective}.
R-Tuning trains models to acknowledge gaps in parametric knowledge \citep{zhang-etal-2024-r}, while \citet{feng-etal-2024-dont} use collaboration between models to detect knowledge gaps and decide whether to abstain.

\paragraph{Abstention benchmarks.}
AbstentionBench includes stale-knowledge questions evaluated without retrieval \citep{kirichenko2025abstentionbench}.
RefusalBench perturbs question--passage pairs to test context-grounded abstention \citep{muhamed2025refusalbench}.
On \name, its context-only answer prompt makes models decline many answerable questions (Appendix~\ref{sec:appendix-external}).
Among its six categories, UAEval4RAG generates out-of-database questions from recent news and evaluates abstention when the database lacks support \citep{peng-etal-2025-unanswerability}.
Our setting permits both retrieved and parametric knowledge and distinguishes acknowledged uncertainty from premise denial, following (QA)$^2$'s distinction between false and unverifiable assumptions \citep{kim-etal-2023-qa}.

\paragraph{Evidence sufficiency in RAG.}
\citet{mallen-etal-2023-trust} show that retrieval helps answer questions about less popular facts that models struggle to recall.
\citet{joren2025sufficient} study whether the retrieved context alone suffices to answer, and how models behave when it does not.
Their evaluation uses answerable questions; \name adds questions designed to expose gaps in both sources.
FLARE retrieves during generation using predicted upcoming sentences \citep{jiang-etal-2023-active}; iterative RAG uses prompted self-assessment of context sufficiency (Appendix~\ref{sec:appendix-retrieval}).
Related but insufficient passages can mislead models \citep{cuconasu2024power}, and prior work studies how further retrieval affects abstention \citep{303de1805f3931282381c71f804bae498f57ade0}.

Appendix~\ref{sec:appendix-related-work} discusses uncertainty estimation, knowledge conflicts, temporal datasets, broader RAG evaluation and Minecraft-based evaluation.

\section{\name: A Dynamic Benchmark for Knowledge-Aware Abstention}
\label{sec:benchmark}
\name evaluates whether a RAG system recognizes when its parametric knowledge and retrieved documents together cannot support an answer.
The questions use real Minecraft content, including additions documented after the corpus cut-off.
When the required facts also postdate the model's knowledge cut-off, the system should acknowledge the missing information.
We distinguish such a response from premise denial, in which the model asserts, without support, that an item or feature named in the question does not exist.
The issue is whether the model has enough knowledge to justify that claim, even if it happens to be correct.
\Cref{tab:example} shows an example of each, alongside a third outcome: an unsupported answer.

\subsection{Temporal Corpora and Knowledge Availability}
We collect two corpora from the \textit{Minecraft} Wiki (\href{https://minecraft.wiki}{\texttt{minecraft.wiki}}), using 31 March 2025 as the corpus cut-off:

\begin{itemize}[leftmargin=*]
    \item \textbf{Pre-cut-off corpus ($C_\text{pre}$):} The latest revision of each wiki page at or before the corpus cut-off: 2,642 pages split into 11,612 chunks as detailed in Appendix~\ref{sec:appendix-scraping}. This is the only corpus available to the evaluated retrieval pipelines. By default they use a \emph{hybrid retriever} that combines keyword (BM25) and embedding similarity (Appendix~\ref{sec:appendix-retrieval}).
    \item \textbf{Post-cut-off corpus ($C_\text{post}$):} Pages created after the corpus cut-off, stored at their latest revision when scraped on 8 September 2026: 170 pages split into 712 chunks. These supply facts for generating unanswerable questions.
\end{itemize}

We exclude pages that existed before the corpus cut-off from $C_\text{post}$, even if they were later updated, to focus on missing knowledge about later additions to the game.
Appendix~\ref{sec:appendix-scraping} provides details on data collection and filtering.

Excluding a fact from retrieval does not by itself exclude it from parametric knowledge.
Temporal separation requires the fact to postdate the model's knowledge cut-off as well as the corpus cut-off.
We report each model's knowledge cut-off when available in \Cref{tab:models}.
Our leakage audit therefore asks models the 300 unanswerable questions without retrieval and scores their answers against reference answers that an oracle model writes from $C_\text{post}$. A model that often answers correctly may have seen the post-cut-off pages in training (\Cref{fig:leakage_model}).

\subsection{Scenario and Question Generation}
\label{sec:scenario-construction}
\Cref{fig:method} traces the generation of one answerable question (top, blue) and one unanswerable question (bottom, orange) through the same five steps.
A generator LLM, \texttt{Qwen3.5 122B-A10B}, performs every writing step, each with its own prompt (Appendix~\ref{sec:appendix-prompts-generation}).
\begin{enumerate}[leftmargin=*]
\item \textbf{Scenario.} We draw three random chunks from $C_\text{pre}$ and give them to the generator. It writes a player \emph{scenario} of under 30 words about one or two related concepts in these chunks.
\item \textbf{Target chunk.} For an answerable question, we embed the scenario with \texttt{Qwen3-Embedding-8B} and retrieve the ten chunks with the most similar embeddings in terms of cosine similarity from $C_\text{pre}$. For an unanswerable question, we follow the same procedure, except the chunks are retrieved from $C_\text{post}$. We drop chunks from technical pages and, for an answerable question, the scenario's own three chunks. The \emph{target chunk} is drawn uniformly at random from the five highest-ranked chunks that remain.
\item \textbf{Fact.} The generator lists up to six facts that appear in the target chunk but not in the comparison text. For an answerable question, the comparison text is the scenario's three chunks, so the scenario does not already contain the fact. For an unanswerable question, it is the three $C_\text{pre}$ chunks most similar to the target chunk, so the fact is new relative to $C_\text{pre}$. One listed fact is chosen uniformly at random.
\item \textbf{Question.} The generator receives the scenario, the fact and the target chunk. It writes one candidate question from the player's perspective whose answer is the fact, and the question is appended to the scenario.
\item \textbf{Verification.} A separate LLM checks each candidate against $C_\text{pre}$ (\Cref{sec:verification}).
\end{enumerate}
Appendix~\ref{sec:appendix-choices} compares \texttt{Qwen3.5 122B-A10B} with two other generators on the same pipeline; we chose to use \texttt{Qwen3.5 122B-A10B} since the resulting answerable and unanswerable questions were the hardest to tell apart by wording alone.
Appendix~\ref{sec:appendix-generation} details the algorithms, and Appendix~\ref{sec:appendix-unasw} shows a generated question of each kind.

Scenarios make the benchmark harder in a realistic way, because they lead retrieval to familiar contexts with insufficient information.
For \DCPremiseHitFive\% of the 300 unanswerable questions, the hybrid retriever returns at least one of the scenario's own source chunks among its five retrieved chunks (\Cref{fig:scenario_passages}).
Such a chunk is related to the scenario but says nothing about the missing fact.
Removing the scenario raises Correct Abstention Rate for all \DFScenModelsStrict{} models tested with single-shot RAG (\Cref{sec:res-scenario}).

\subsection{Verification and Benchmark Assembly}
\label{sec:verification}
Generation can go wrong in three ways:
\begin{itemize}[leftmargin=*]
\item An answerable question whose answer is not in $C_\text{pre}$;
\item An unanswerable question whose fact is already in $C_\text{pre}$, for example because a page created after the corpus cut-off restates an older fact;
\item Many questions drawn from the same few pages.
\end{itemize}
The first two mislabel a question, and the third lets a few pages dominate the benchmark. Three steps remove them.

\paragraph{Automated verification.} A separate verifier LLM, \texttt{Gemma 4 31B}, checks each candidate scenario and question, so that no model judges its own questions.
For each query, the verifier reads the 10 $C_\text{pre}$ chunks retrieved with hybrid retrieval and reranking.
\emph{Reranking} uses a second model to reorder the retrieved chunks by their relevance to the query.
An answerable candidate is kept only if both of the following hold: (1) the scenario does not state or directly imply the target fact, so it does not give the answer away; and (2) the chunks retrieved for the question contain the information needed to answer it.
An unanswerable candidate is kept only if both of the following hold: (1) the $C_\text{pre}$ chunks retrieved for the target fact do not state or directly imply it; and (2) the chunks retrieved for the question do not answer it.
Appendix~\ref{sec:appendix-prompts} gives the verifier's prompts.

\paragraph{Manual review.}
We reviewed all unanswerable questions against $C_\text{pre}$ and dropped 21 malformed candidates (claiming fake in-game items like ``steel'' and fabricated game mechanics like ``attacked by a creeper with a diamond sword''). We found no examples of questions which had answers in $C_\text{pre}$. We keep questions that can be answered only by assuming that an older mechanic carries over to a new item, such as whether new copper tools accept an enchantment that iron tools already accept: such a guess may be right, but neither $C_\text{pre}$ nor the model's training establishes it, so the model should still abstain.

\paragraph{Assembly.}
The number of chunks per $C_\text{post}$ page varies widely (median \DMPostChunksMedian{}, up to \DMPostChunksMax{}), so without limits the questions concentrate on a few pages. We therefore allow at most 2 questions per chunk and at most 10 per page.
We also drop candidates whose question sentence has a novelty cue (``new,'' ``recently'', a month or a year), which would reveal a post-cut-off subject.

The final benchmark contains 700 answerable and 300 unanswerable questions.
For the 700 answerable questions, the default hybrid retriever returns the target chunk among its five retrieved chunks in \DCRecallHitFive\% of cases.
Answerability in the corpus therefore does not guarantee that the LLM receives the target passage.

\section{Results}
\label{sec:results}
In what follows, we introduce our results over \DFNModels{} LLMs evaluated on \name without retrieval, with single-shot RAG and with iterative RAG. We assess whether models acknowledge missing knowledge on unanswerable questions while attempting and correctly answering answerable questions. After describing the evaluation setup and metrics (\Cref{sec:metrics}), we compare retrieval conditions (\Cref{sec:retrieval-results}), examine the associations of Correct Abstention Rate with model size and release date (\Cref{sec:res-size}), and measure the effect of enabling thinking (\Cref{sec:res-thinking}). Throughout these results, we distinguish acknowledged uncertainty from premise denial. Appendix~\ref{sec:appendix-results} reports the results for every model and further analyses.

\subsection{Evaluation Setup and Metrics}
\label{sec:metrics}
We use the 700 answerable and 300 unanswerable questions constructed in \Cref{sec:benchmark}. A \emph{configuration} is one model with one thinking setting. \DFThinkModels{} models have \emph{switchable} thinking (the same weights run with thinking enabled or disabled); running both gives \DFRagPairs{} configurations per retrieval condition. Appendix~\ref{sec:appendix-setup} gives the API settings and statistical tests; Appendix~\ref{sec:appendix-models} lists the models and their knowledge cut-offs, and \Cref{tab:results_rag0,tab:results_rag2} report every model and thinking setting under each condition.

\textbf{Correct Abstention Rate} (CAR) is the percentage of unanswerable questions on which the model acknowledges insufficient knowledge without a premise denial or an unsupported answer. An LLM \emph{response judge} labels each response as an abstention, a premise denial or an attempted answer (Appendix~\ref{sec:appendix-judges}). A premise denial followed by a guessed answer still counts as a premise denial, not a correct abstention. Premise denials of answerable questions count as attempts. Appendix~\ref{sec:appendix-judges} repeats the analyses under a lenient rule that also counts premise denials as abstentions: the retrieval findings hold, whereas the thinking and release-date findings change. \textbf{Answerable Attempt Rate} is the percentage of answerable questions the model attempts to answer. \textbf{Answer Accuracy} averages the scores of three LLM \emph{accuracy judges} from different families on the answerable questions: 1 for correct, 0.5 for partly correct and 0 for incorrect answers or abstentions, against \emph{reference answers} that an \emph{oracle model} writes from each question's source pages. Outputs that exhaust the output budget of 16,000 tokens, including thinking, without a final answer (\DCExhaustedShare\% of all answers; Appendix~\ref{sec:appendix-setup}) are excluded from the two rates and score 0 in Answer Accuracy. Appendix~\ref{sec:appendix-judges} details the scoring.

\paragraph{Human validation.} Three annotators independently label \HAOneN{} responses and score \HATwoN{} attempted answers for accuracy. The response judge matches the human majority on \HAOneMajAgree\% of response labels; disagreements mainly concern an abstention sentence followed by a premise denial. For accuracy, mean judge--human agreement is $\kappa=\HATwoJudgeKappa$, versus $\HATwoHumanKappa$ between humans (quadratic-weighted Cohen's $\kappa$). Samples are balanced across response labels and accuracy-score bins; Appendix~\ref{sec:appendix-human} gives details and confidence intervals.

\begin{figure}[!t]
    \centering
    \includegraphics[width=\columnwidth]{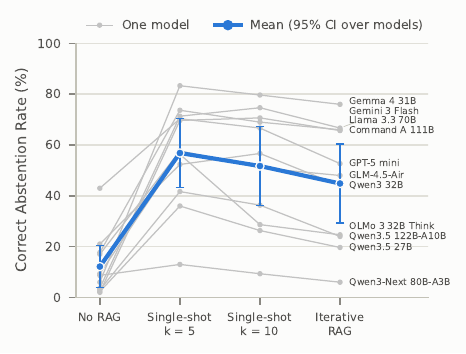}
    \caption{Retrieval-depth ablation on \DFDepthModels{} models with a $k{=}10$ run. $k$ is the number of chunks retrieved per query. Thinking is disabled where switchable; other models use their defaults. Gray lines: individual models; blue: mean Correct Abstention Rate with a 95\% confidence interval over models. Appendix~\ref{sec:res-depth} gives the comparisons and their limits.}
    \label{fig:depth}
\end{figure}

\begin{figure}[!t]
    \centering
    \includegraphics[width=\columnwidth]{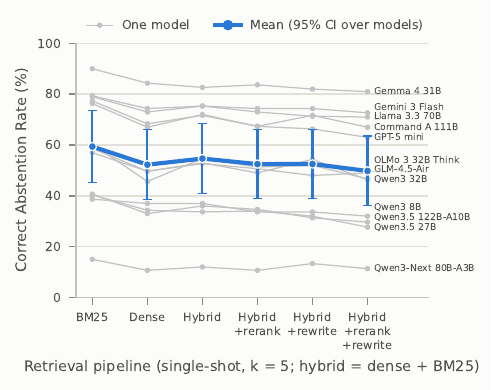}
    \caption{Retrieval-pipeline ablation on \DFPipeModels{} models, with single-shot RAG and five chunks. The six pipelines are keyword (BM25) or embedding (dense) search alone, the default hybrid of the two, and the hybrid with reranking, query rewriting or both (Appendix~\ref{sec:appendix-retrieval}). Thinking is disabled where switchable; other models use their defaults. Gray lines: individual models; blue: mean Correct Abstention Rate with a 95\% confidence interval over models. Appendix~\ref{sec:res-pipeline} reports the pipeline comparisons.}
    \label{fig:pipeline}
\end{figure}

\subsection{Retrieval Raises Correct Abstention Rate; Iterative RAG Lowers It Again}
\label{sec:retrieval-results}
Single-shot RAG retrieves five chunks from $C_\text{pre}$ using the default hybrid retriever (BM25 and dense retrieval, combined by reciprocal rank fusion; see details in Appendix~\ref{sec:appendix-retrieval}). \Cref{fig:depth} plots Correct Abstention Rate (y-axis) against the retrieval condition (x-axis) for the subset of \DFDepthModels{} models also evaluated with ten retrieved chunks; gray lines show individual models, and the blue line shows their mean. Across all \DFRagPairs{} configurations, mean Correct Abstention Rate rises from \DFMeanNoRag\% without retrieval to \DFMeanRag\% with single-shot RAG. Relative to no retrieval, \DFRagUp{} configurations show a statistically significant increase in Correct Abstention Rate, and none shows a statistically significant decrease ($q<0.05$; see Appendix~\ref{sec:appendix-stats} for details on statistical analysis).

Our iterative RAG implementation uses prompted self-assessment, inspired by Self-RAG~\citep{asai2023selfrag} and query-level self-evaluation~\citep{303de1805f3931282381c71f804bae498f57ade0}. In our experiments, it starts with the same five chunks and lets the model request up to two additional retrieval rounds when it judges the context insufficient (Algorithm~\ref{alg:iterative}; Appendix~\ref{sec:appendix-iterative-rag}). Relative to single-shot RAG, mean Correct Abstention Rate falls to \DFMeanIter\%, with significant decreases in \DFItDown{} of the same \DFItPairs{} configurations and no significant increases. Mean Answer Accuracy on the answerable questions changes from \DFAccMeanRag\% to \DFAccMeanIter\%, compared with \DFAccMeanNoRag\% without retrieval. \Cref{fig:retrieval} in the appendix shows both metrics for each configuration. Thus, in this pipeline, further retrieval lowers CAR while adding little accuracy.

The Sulfur Cube example in Appendix~\ref{sec:qualitative} illustrates this failure: \texttt{GPT-OSS 120B} correctly abstains with single-shot RAG, then uses a retrieved passage about Magma Cubes to answer about the unfamiliar entity after further retrieval. The example shows how relevant-looking context can support an unwarranted substitution; the aggregate comparison does not isolate which component of iterative RAG causes the decrease, and the reflection prompt itself asks whether the context suffices to answer rather than whether to answer at all.

Appendix~\ref{sec:res-depth} examines how retrieving five versus ten chunks affects Correct Abstention Rate for \DFDepthModels{} models (\Cref{fig:depth}) and discusses why iterative RAG cannot isolate the effect of follow-up queries from that of additional context. Appendix~\ref{sec:res-pipeline} compares six retrieval pipelines (\Cref{fig:pipeline}): adding both query rewriting and reranking to hybrid retrieval lowers Correct Abstention Rate for all \DFPipeFullBelow{} tested models. Appendix~\ref{sec:res-scenario} details the scenario-removal ablation introduced in \Cref{sec:scenario-construction}: removing scenarios from both retrieval queries and model inputs raises Correct Abstention Rate for all \DFScenQonlyHigherStrict{} tested models under single-shot RAG.

\begin{figure}[!t]
    \centering
    \includegraphics[width=\columnwidth]{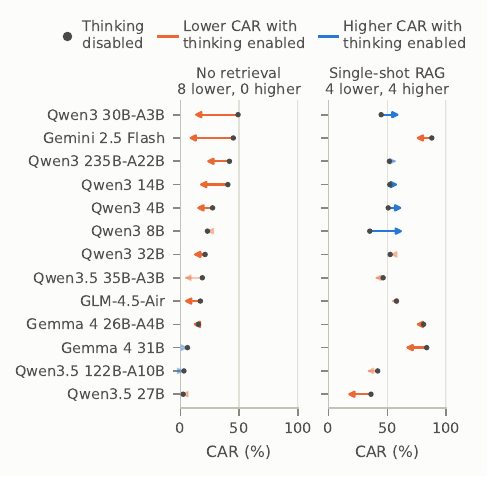}
    \caption{Correct Abstention Rate with thinking disabled (gray dot) and enabled (arrowhead), holding weights fixed for the \DFThinkModels{} models with switchable thinking. Orange: lower with thinking enabled; blue: higher; faded: not significant ($q\geq0.05$).}
    \label{fig:thinking}
\end{figure}

\begin{figure}[!t]
    \centering
    \includegraphics[width=\columnwidth]{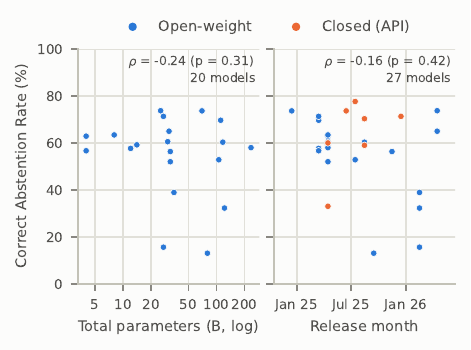}
    \caption{Correct Abstention Rate versus total parameters (\DFNTotal{} open-weight models, left) and release month (\DFNReleaseModel{} models, right; closed models in orange), with single-shot RAG. One point per model, with thinking enabled where switchable; neither Spearman correlation is statistically significant at the 5\% level. \Cref{fig:size_active} in the appendix shows active parameters.}
    \label{fig:size}
\end{figure}

\begin{figure*}[!t]
    \begin{minipage}[t]{\columnwidth}
        \centering
        \includegraphics[width=\linewidth]{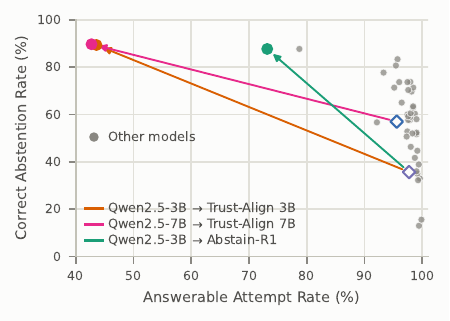}
        \caption{Answerable Attempt Rate against Correct Abstention Rate, single-shot RAG. Gray: the \DAMainN{} configurations of the other models. Arrows: baseline model (hollow) to fine-tuned model (filled); \texttt{Abstain-R1} also changes the prompt template.}
        \label{fig:abstention_tradeoff}
    \end{minipage}\hfill
    \begin{minipage}[t]{\columnwidth}
        \centering
        \includegraphics[width=\linewidth]{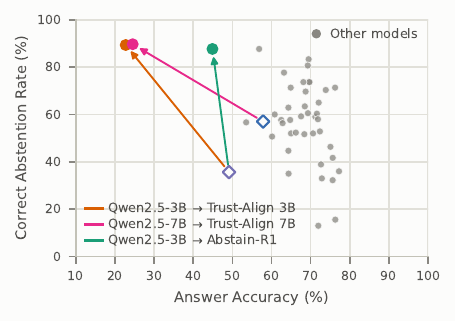}
        \caption{Answer Accuracy against Correct Abstention Rate with single-shot RAG, for the same models as \Cref{fig:abstention_tradeoff}. Gray: the \DAMainN{} configurations of the other models. Arrows: baseline model (hollow) to fine-tuned model (filled).}
        \label{fig:abstention_accuracy}
    \end{minipage}
\end{figure*}

\subsection{Correct Abstention Rate Has No Significant Association with Size or Release Date}
\label{sec:res-size}
Under single-shot RAG, neither total parameters (Spearman $\rho=\DFRhoTotal$, $p=\DFPTotal$) nor active parameters ($\rho=\DFRhoActive$, $p=\DFPActive$) has a significant association with Correct Abstention Rate across the \DFNTotal{} open-weight models. Active parameters are those used for each token. A mixture-of-experts model uses only part of its total parameters for each token. \Cref{fig:size} (left) shows the comparison with total parameters, with thinking enabled where switchable; \Cref{fig:size_active} in the appendix shows active parameters. Release month likewise has no significant association across all \DFNReleaseModel{} models ($\rho=\DFRhoReleaseModel$, $p=\DFPReleaseModel$; \Cref{fig:size}, right), or across the \DFNReleaseFamily{} families when selecting the model with the highest Correct Abstention Rate in each ($\rho=\DFRhoReleaseFamily$, $p=\DFPReleaseFamily$). On the same open-weight models, Answer Accuracy is positively associated with total parameters ($\rho=\DFRhoAccTotal$, $p=\DFPAccTotal$; \Cref{fig:size_accuracy} in the appendix). Thus, the observed accuracy gains with scale do not establish better recognition of missing knowledge.

The response categories matter for interpreting this result. Recent families such as Qwen3.5 and Qwen3-Next frequently respond with premise denial: they assert, without support, that the queried item does not exist. \Cref{fig:outcomes,fig:outcomes_rag} in the appendix separate premise denials from abstentions and attempted answers, by family and model. As the copper horse armor responses in \Cref{tab:example} illustrate, a premise denial can also lead to a numerical answer or the substitution of a familiar item. Such a response does not establish recognition of a knowledge gap, even if its claim was true at the corpus cut-off.

\subsection{Thinking Lowers Correct Abstention Rate Without Retrieval}
\label{sec:res-thinking}
\Cref{fig:thinking} compares thinking disabled (gray dots) and enabled (arrowheads) for the \DFThinkModels{} models with switchable thinking, holding weights fixed, without retrieval (left) and with single-shot RAG (right). Orange indicates lower CAR, blue higher; faded arrows mark changes that are not statistically significant.

Without retrieval, we observe \DFThinkNoRagDown{} statistically significant decreases in CAR and no statistically significant increases. Per question, enabling thinking without retrieval turns abstentions into premise denials more often than the reverse for \DTNoRagSigAtoD{} models (\Cref{fig:thinking_pairs}). Appendix~\ref{sec:res-thinking-details} shows that Answer Accuracy can improve while CAR falls.

Across \DFNAccAbst{} single-shot RAG configurations, Answer Accuracy and CAR have no statistically significant unadjusted association (Spearman $\rho=\DFRhoAccAbst$, $p=\DFPAccAbst$); Appendix~\ref{sec:res-accuracy} reports a negative association after adjusting for release date.

With single-shot RAG, \DFTradeBothUp{} of \DFRagPairs{} configurations have both a higher CAR and a higher Answerable Attempt Rate than without retrieval (Appendix~\ref{sec:res-tradeoff}).

\subsection{Models Fine-Tuned to Abstain}
\label{sec:res-abstention}
Some models are fine-tuned specifically for abstention, such as \texttt{Trust-Align 3B}, \texttt{Trust-Align 7B}~\citep{song2025trustalign}, and \texttt{Abstain-R1 3B}~\citep{zhai2026abstainr1}. We evaluate these three models against their corresponding base model (Appendix~\ref{sec:res-abstention-trained}). \Cref{fig:abstention_tradeoff} shows that fine-tuning raises CAR and lowers Answerable Attempt Rate. With single-shot RAG, fine-tuning raises CAR from \DAQThreeRagCar\% (\texttt{Qwen2.5-3B-Instruct}) to \DATAThreeRagCar\% (\texttt{Trust-Align 3B}), but \texttt{Trust-Align 3B} attempts only \DATAThreeRagAtt\% of answerable questions. These models lie on or near the Pareto trade-off frontier of the other configurations, the highest CAR that they reach at or above each Answerable Attempt Rate. They trade a higher CAR for lower Answerable Attempt Rate and Answer Accuracy by abstaining on answerable questions that other models answer correctly (\Cref{fig:abstention_accuracy}). Fine-tuning moves these three models along the trade-off frontier rather than beyond it. 

\section{Conclusion}

We introduced \name, a dynamic benchmark built from dated \textit{Minecraft} Wiki pages to evaluate recognition of missing knowledge and distinguish acknowledged uncertainty from premise denial. Across \DFNModels{} LLMs, single-shot RAG produces statistically significant increases in Correct Abstention Rate in \DFRagUp{} of \DFRagPairs{} configurations relative to no retrieval; our prompted iterative RAG pipeline produces statistically significant decreases in \DFItDown{} of the same configurations relative to single-shot RAG, with little change in Answer Accuracy. Under single-shot RAG, Correct Abstention Rate has no statistically significant association with model size or release date. Enabling thinking produces statistically significant decreases in Correct Abstention Rate for \DFThinkNoRagDown{} of \DFThinkModels{} models without retrieval, with no statistically significant increases; its effect with single-shot RAG varies across models.

These results motivate evaluating and training recognition of missing knowledge explicitly. A premise denial does not establish this recognition when the model cannot verify its claim. Future work should improve judgments of evidence sufficiency across retrieval rounds and test whether these findings extend to other domains with dated documentation.

\section{Limitations} %
While \name provides a controlled environment to evaluate the temporal adaptability and abstention capabilities of RAG-equipped LLMs, we identify several limitations to our work.

\subsection{Domain Concentration of the Corpus}
Our framework uses only the \textit{Minecraft} Wiki, which may not represent the complexity of other domains where RAG systems are deployed. Other domains may have more erroneous chunks or poorly maintained version histories, or may not receive regular updates. Future work is needed to test whether our findings on knowledge unavailability at a corpus cut-off generalize to other domains.

\subsection{Dependence on LLM Judges}
We rely on LLMs throughout: \texttt{Qwen3.5 122B-A10B} writes the questions, \texttt{Gemma 4 31B} verifies them, the response judge \texttt{Gemma 4 26B-A4B} labels abstentions and premise denials, the oracle model \texttt{DeepSeek V4.1 Flash} writes reference answers, and three accuracy judges score the answers. These models may impart their own biases, such as preferring longer answers or asking underspecified questions. A human study on 100 shared items (\Cref{sec:appendix-human}) finds that the judges agree with the annotators at least as well as the annotators agree with one another, but it covers a small sample. The two accuracy judges whose families we also evaluate show a small self-preference (\Cref{sec:res-judge-selfpref}).

\subsection{One Chunking Scheme and Embedding Model}
We compare six retrieval pipelines, including reranking and query rewriting, but all of them search the same index: fixed-size chunks of about 1,500 characters split at line breaks, embedded with \texttt{Qwen3-Embedding-8B}. Hierarchical or semantic chunking, or other embedding models, may retrieve differently.

\subsection{Scoring Premise Denials}
Our headline Correct Abstention Rate excludes premise denials, including premise denials followed by a guessed answer. We report these responses separately from acknowledged uncertainty (Appendix~\ref{sec:appendix-judges}). The answer prompt did not tell models that the game may contain items newer than their knowledge. We have not tested whether making this temporal uncertainty explicit changes the frequency of premise denials.

\subsection{Limited Evaluation Models}
Because of cloud inference costs and limited local hardware, we selected models to cover a wide range of sizes, architectures and families. We chose the Qwen3 family because it offers many sizes and architectures within one family, which lets us vary one factor while holding the others fixed.

\section*{Acknowledgments}
This work was funded in part by the National Science Foundation (NSF) award CNS-2504819 and the Lilly Endowment. Any opinions, findings and conclusions or recommendations expressed in this material are those of the authors and do not necessarily reflect the views of the sponsors.

\bibliography{custom}

\appendix
\renewcommand{\floatpagefraction}{0.7}\renewcommand{\dblfloatpagefraction}{0.85}
\renewcommand{\topfraction}{0.9}\renewcommand{\dbltopfraction}{0.9}\renewcommand{\textfraction}{0.07}
\let\DCappendixsection\section
\renewcommand{\section}{\FloatBarrier\DCappendixsection}

\section{Additional Results}
\label{sec:appendix-results}

This section collects the figures and tables supporting the claims in the main text.

\paragraph{Results per model.} \Cref{tab:results_rag0} reports single-shot RAG results and \Cref{tab:results_rag2} all three retrieval conditions, for every configuration.

\begin{table*}[!tp]
\centering
\scriptsize
\begin{tabular}{lrrrrr}
\toprule
Model & Attempt \% & CAR \% & Accuracy \% & \shortstack{Accuracy when\\attempted \%} & Bal. F1 \\
\midrule
Gemma 4 26B-A4B\textsuperscript{R} & 95.4 [93.6, 96.7] & 80.7 [75.8, 84.7] & 69.3 [67.2, 71.4] & 72.6 & 88.9 \\
Gemma 4 26B-A4B\textsuperscript{R} (thinking) & 97.9 [96.6, 98.8] & 73.7 [68.3, 78.6] & 69.6 [67.5, 71.7] & 73.6 & 87.4 \\
Gemma 4 31B\textsuperscript{V} & 95.7 [93.9, 97.0] & 83.3 [78.7, 87.1] & 69.5 [67.5, 71.6] & 72.6 & \textbf{90.1} \\
Gemma 4 31B\textsuperscript{V} (thinking) & 96.4 [94.8, 97.6] & 65.0 [59.4, 70.2] & 72.0 [70.1, 74.0] & 74.7 & 83.3 \\
\addlinespace[2pt]
Gemma 3 4B & 92.1 [89.9, 93.9] & 56.7 [51.0, 62.2] & 53.5 [51.0, 56.0] & 58.1 & 78.3 \\
Gemma 3 12B & 97.9 [96.5, 98.7] & 57.7 [52.0, 63.1] & 62.5 [60.2, 64.7] & 63.8 & 81.5 \\
Gemma 3 27B & 98.4 [97.2, 99.1] & 71.3 [66.0, 76.2] & 64.9 [62.6, 67.0] & 65.9 & 86.7 \\
\addlinespace[2pt]
Llama 3.3 70B & 97.4 [96.0, 98.4] & 73.7 [68.4, 78.3] & 68.1 [66.2, 70.3] & 69.9 & 87.1 \\
\addlinespace[2pt]
Command A 111B & 98.1 [96.8, 98.9] & 69.7 [64.2, 74.6] & 68.7 [66.5, 71.0] & 70.0 & 85.9 \\
\addlinespace[2pt]
GLM-4.5-Air & 97.6 [96.1, 98.5] & 57.7 [52.0, 63.1] & 64.8 [62.5, 66.9] & 66.4 & 81.3 \\
GLM-4.5-Air (thinking) & 97.4 [95.9, 98.4] & 52.8 [47.2, 58.4] & 72.4 [70.1, 74.5] & 74.8 & 79.7 \\
\addlinespace[2pt]
OLMo 3 32B Think & 95.7 [93.9, 97.0] & 56.3 [50.7, 61.8] & 62.8 [60.4, 65.3] & 65.6 & 80.0 \\
\addlinespace[2pt]
Qwen3 4B & 97.3 [95.8, 98.3] & 50.7 [45.0, 56.3] & 60.1 [57.6, 62.5] & 61.8 & 78.9 \\
Qwen3 4B (thinking) & 98.4 [97.2, 99.1] & 62.9 [57.2, 68.2] & 64.3 [61.8, 66.5] & 66.5 & 83.6 \\
Qwen3 8B & 99.1 [98.1, 99.6] & 35.0 [29.8, 40.6] & 64.4 [62.3, 66.7] & 64.9 & 75.1 \\
Qwen3 8B (thinking) & 98.4 [97.2, 99.1] & 63.4 [57.8, 68.7] & 68.5 [66.3, 70.7] & 70.4 & 83.8 \\
Qwen3 14B & 98.9 [97.8, 99.4] & 52.0 [46.4, 57.6] & 65.0 [62.7, 67.1] & 65.7 & 80.1 \\
Qwen3 14B (thinking) & 97.6 [96.1, 98.5] & 59.2 [53.5, 64.6] & 67.5 [65.3, 69.8] & 69.4 & 81.9 \\
Qwen3 30B-A3B & 99.1 [98.1, 99.6] & 44.7 [39.1, 50.3] & 64.3 [62.1, 66.4] & 64.9 & 77.9 \\
Qwen3 30B-A3B (thinking) & 98.0 [96.7, 98.8] & 60.5 [54.9, 65.9] & 69.3 [67.0, 71.5] & 71.0 & 82.5 \\
Qwen3 32B & 99.0 [98.0, 99.5] & 52.3 [46.7, 57.9] & 66.2 [64.0, 68.4] & 66.9 & 80.3 \\
Qwen3 32B (thinking) & 98.3 [97.0, 99.0] & 52.0 [46.4, 57.6] & 70.6 [68.3, 72.7] & 72.1 & 79.8 \\
Qwen3 235B-A22B & 99.0 [98.0, 99.5] & 51.7 [46.0, 57.3] & 68.4 [66.2, 70.4] & 69.1 & 80.1 \\
Qwen3 235B-A22B (thinking) & 99.0 [98.0, 99.5] & 58.0 [52.3, 63.4] & 71.9 [69.9, 74.0] & 72.6 & 82.2 \\
\addlinespace[2pt]
Qwen3-Next 80B-A3B & 99.4 [98.5, 99.8] & 13.0 [9.7, 17.3] & 72.0 [69.6, 74.1] & 72.4 & 69.4 \\
\addlinespace[2pt]
Qwen3.5 27B & 99.0 [98.0, 99.5] & 36.0 [30.8, 41.6] & \textbf{77.2} [75.3, 79.1] & 78.0 & 75.3 \\
Qwen3.5 27B (thinking) & \textbf{99.9} [99.2, 100.0] & 15.6 [11.6, 20.7] & 76.3 [74.2, 78.3] & 78.6 & 70.2 \\
Qwen3.5 35B-A3B & 98.0 [96.7, 98.8] & 46.3 [40.8, 52.0] & 75.0 [73.0, 77.0] & 76.6 & 77.9 \\
Qwen3.5 35B-A3B (thinking) & 99.4 [98.4, 99.8] & 38.9 [31.7, 46.6] & 72.6 [70.6, 74.8] & 77.8 & 76.3 \\
Qwen3.5 122B-A10B\textsuperscript{G} & 98.7 [97.6, 99.3] & 41.7 [36.2, 47.3] & 75.6 [73.5, 77.8] & 76.6 & 76.8 \\
Qwen3.5 122B-A10B\textsuperscript{G} (thinking) & 99.3 [98.3, 99.7] & 32.2 [26.3, 38.8] & 75.6 [73.7, 77.6] & 77.8 & 74.4 \\
\addlinespace[2pt]
GPT-4.1 mini & 99.6 [98.7, 99.9] & 33.0 [27.9, 38.5] & 72.9 [70.8, 74.9] & 73.2 & 74.7 \\
GPT-4.1 nano & 97.4 [96.0, 98.4] & 60.0 [54.4, 65.4] & 60.8 [58.4, 63.2] & 62.4 & 82.1 \\
\addlinespace[2pt]
GPT-5 mini & 97.6 [96.1, 98.5] & 70.3 [64.9, 75.2] & 73.9 [71.7, 76.0] & 75.7 & 85.9 \\
GPT-5 nano & 98.0 [96.7, 98.8] & 59.0 [53.4, 64.4] & 71.2 [69.0, 73.4] & 72.6 & 82.0 \\
\addlinespace[2pt]
GPT-OSS 120B & 98.7 [97.6, 99.3] & 60.3 [54.7, 65.7] & 71.6 [69.4, 73.6] & 72.5 & 82.8 \\
\addlinespace[2pt]
Gemini 2.5 Flash & 78.7 [75.5, 81.6] & \textbf{87.7} [83.5, 90.9] & 56.8 [54.0, 59.7] & 72.2 & 82.4 \\
Gemini 2.5 Flash (thinking) & 96.0 [94.3, 97.2] & 73.7 [68.4, 78.3] & 69.7 [67.5, 71.7] & 72.6 & 86.4 \\
Gemini 2.5 Flash-Lite & 93.3 [91.2, 94.9] & 77.7 [72.6, 82.0] & 63.2 [61.0, 65.5] & 67.7 & 86.5 \\
\addlinespace[2pt]
Gemini 3 Flash & 95.1 [93.3, 96.5] & 71.3 [66.0, 76.2] & 76.2 [74.3, 78.3] & \textbf{80.1} & 85.0 \\
\bottomrule
\end{tabular}
\caption{Results with single-shot RAG (hybrid retrieval, $k{=}5$): one row per model, with a second row, marked ``(thinking)'', for models with switchable thinking. \emph{Attempt}: Answerable Attempt Rate; \emph{CAR}: Correct Abstention Rate; \emph{Accuracy when attempted}: Answer Accuracy over the answerable questions the model attempted; \emph{Bal.\ F1}: balanced F1, defined in \Cref{sec:appendix-class-balance}. Brackets are 95\% intervals (Wilson for rates, bootstrap over questions for accuracy). The best value in each column is bold. Superscripts mark models with a role in building or scoring the benchmark (G generator, V verifier and accuracy judge, R response judge).}
\label{tab:results_rag0}
\end{table*}

\begin{table*}[!tp]
\centering
\scriptsize
\begin{tabular}{lrrrrrrrrr}
\toprule
 & \multicolumn{3}{c}{CAR \%} & \multicolumn{3}{c}{Answerable Attempt \%} & \multicolumn{3}{c}{Answer Accuracy \%} \\
\cmidrule(lr){2-4}\cmidrule(lr){5-7}\cmidrule(lr){8-10}
Model & No RAG & RAG & Iter. & No RAG & RAG & Iter. & No RAG & RAG & Iter. \\
\midrule
Gemma 4 26B-A4B\textsuperscript{R} & 15.3 & 80.7 & 67.7 & 88.9 & 95.4 & 96.4 & 26.9 & 69.3 & 69.7 \\
Gemma 4 26B-A4B\textsuperscript{R} (thinking) & 10.2 & 73.7 & 71.2 & 90.8 & 97.9 & 98.2 & 26.6 & 69.6 & 70.6 \\
Gemma 4 31B\textsuperscript{V} & 6.0 & 83.3 & 76.0 & 95.0 & 95.7 & 97.3 & 33.4 & 69.5 & 71.2 \\
Gemma 4 31B\textsuperscript{V} (thinking) & 6.7 & 65.0 & 60.7 & 93.2 & 96.4 & 97.7 & 34.0 & 72.0 & 74.0 \\
\addlinespace[2pt]
Gemma 3 4B & 25.7 & 56.7 & 38.7 & 90.3 & 92.1 & 92.9 & 14.7 & 53.5 & 53.2 \\
Gemma 3 12B & 9.0 & 57.7 & 42.3 & 96.3 & 97.9 & 98.9 & 18.1 & 62.5 & 63.2 \\
Gemma 3 27B & 14.0 & 71.3 & 61.0 & 94.6 & 98.4 & 98.0 & 20.6 & 64.9 & 64.9 \\
\addlinespace[2pt]
Llama 3.3 70B & 17.7 & 73.7 & 66.0 & 94.9 & 97.4 & 97.9 & 26.5 & 68.1 & 68.1 \\
\addlinespace[2pt]
Command A 111B & 2.0 & 69.7 & 65.7 & 98.6 & 98.1 & 97.9 & 26.0 & 68.7 & 70.0 \\
\addlinespace[2pt]
GLM-4.5-Air & 17.0 & 57.7 & 48.0 & 97.7 & 97.6 & 97.9 & 26.4 & 64.8 & 65.6 \\
GLM-4.5-Air (thinking) & 2.8 & 52.8 & 40.3 & 97.0 & 97.4 & 98.3 & 27.7 & 72.4 & 72.8 \\
\addlinespace[2pt]
OLMo 3 32B Think & 9.3 & 56.3 & 24.7 & 96.3 & 95.7 & 98.0 & 17.3 & 62.8 & 64.4 \\
\addlinespace[2pt]
Qwen3 4B & 27.3 & 50.7 & 40.0 & 84.3 & 97.3 & 97.4 & 12.3 & 60.1 & 60.6 \\
Qwen3 4B (thinking) & 13.2 & 62.9 & 51.5 & 89.6 & 98.4 & 99.1 & 12.0 & 64.3 & 65.5 \\
Qwen3 8B & 23.0 & 35.0 & 33.3 & 92.3 & 99.1 & 99.7 & 17.1 & 64.4 & 64.6 \\
Qwen3 8B (thinking) & 22.2 & 63.4 & 52.7 & 85.3 & 98.4 & 99.0 & 12.2 & 68.5 & 68.2 \\
Qwen3 14B & 40.3 & 52.0 & 49.7 & 83.9 & 98.9 & 99.3 & 17.6 & 65.0 & 66.1 \\
Qwen3 14B (thinking) & 15.6 & 59.2 & 53.4 & 86.5 & 97.6 & 98.6 & 16.5 & 67.5 & 69.0 \\
Qwen3 30B-A3B & 49.0 & 44.7 & 38.3 & 76.6 & 99.1 & 99.1 & 16.3 & 64.3 & 65.0 \\
Qwen3 30B-A3B (thinking) & 11.4 & 60.5 & 51.7 & 88.5 & 98.0 & 97.9 & 18.4 & 69.3 & 69.7 \\
Qwen3 32B & 21.0 & 52.3 & 44.0 & 95.3 & 99.0 & 99.0 & 20.6 & 66.2 & 68.3 \\
Qwen3 32B (thinking) & 10.5 & 52.0 & 45.8 & 96.1 & 98.3 & 98.6 & 20.3 & 70.6 & 70.7 \\
Qwen3 235B-A22B & 41.7 & 51.7 & 45.7 & 86.9 & 99.0 & \textbf{99.9} & 24.1 & 68.4 & 68.9 \\
Qwen3 235B-A22B (thinking) & 21.7 & 58.0 & 45.3 & 86.9 & 99.0 & 99.0 & 23.3 & 71.9 & 72.2 \\
\addlinespace[2pt]
Qwen3-Next 80B-A3B & 8.7 & 13.0 & 6.0 & 91.7 & 99.4 & 99.3 & 22.5 & 72.0 & 72.4 \\
\addlinespace[2pt]
Qwen3.5 27B & 2.3 & 36.0 & 19.7 & 97.3 & 99.0 & 99.6 & 22.4 & \textbf{77.2} & 78.5 \\
Qwen3.5 27B (thinking) & 0.5 & 15.6 & 12.8 & 98.2 & \textbf{99.9} & \textbf{99.9} & 26.3 & 76.3 & \textbf{79.2} \\
Qwen3.5 35B-A3B & 18.7 & 46.3 & 37.3 & 80.0 & 98.0 & 99.0 & 20.1 & 75.0 & 75.2 \\
Qwen3.5 35B-A3B (thinking) & 3.1 & 38.9 & 21.1 & 96.5 & 99.4 & 99.5 & 18.3 & 72.6 & 71.6 \\
Qwen3.5 122B-A10B\textsuperscript{G} & 3.0 & 41.7 & 24.0 & 98.1 & 98.7 & 99.1 & 28.4 & 75.6 & 77.1 \\
Qwen3.5 122B-A10B\textsuperscript{G} (thinking) & 3.5 & 32.2 & 23.1 & 96.7 & 99.3 & 99.6 & 29.3 & 75.6 & 77.1 \\
\addlinespace[2pt]
GPT-4.1 mini & 0.0 & 33.0 & 24.7 & \textbf{99.9} & 99.6 & 99.7 & 35.0 & 72.9 & 73.3 \\
GPT-4.1 nano & 11.0 & 60.0 & 51.3 & 95.9 & 97.4 & 97.7 & 23.6 & 60.8 & 60.7 \\
\addlinespace[2pt]
GPT-5 mini & 43.0 & 70.3 & 52.7 & 76.0 & 97.6 & 99.1 & 37.7 & 73.9 & 76.9 \\
GPT-5 nano & 42.0 & 59.0 & 42.0 & 71.7 & 98.0 & 99.0 & 28.3 & 71.2 & 70.6 \\
\addlinespace[2pt]
GPT-OSS 120B & 2.0 & 60.3 & 45.0 & 99.4 & 98.7 & 99.4 & 31.4 & 71.6 & 73.7 \\
\addlinespace[2pt]
Gemini 2.5 Flash & 45.0 & \textbf{87.7} & \textbf{84.0} & 69.9 & 78.7 & 82.0 & 23.8 & 56.8 & 59.4 \\
Gemini 2.5 Flash (thinking) & 6.7 & 73.7 & 62.0 & 95.9 & 96.0 & 96.4 & 38.5 & 69.7 & 70.8 \\
Gemini 2.5 Flash-Lite & \textbf{56.7} & 77.7 & 70.0 & 74.7 & 93.3 & 94.0 & 23.9 & 63.2 & 63.6 \\
\addlinespace[2pt]
Gemini 3 Flash & 3.7 & 71.3 & 66.7 & 97.6 & 95.1 & 96.7 & \textbf{54.1} & 76.2 & 76.4 \\
\bottomrule
\end{tabular}
\caption{Each model and thinking setting under the three retrieval conditions: Correct Abstention Rate, Answerable Attempt Rate and Answer Accuracy without retrieval, with single-shot RAG and with iterative RAG. Rows and marks as in \Cref{tab:results_rag0}. The best value in each column is bold.}
\label{tab:results_rag2}
\end{table*}

\begin{figure}[!htbp]
    \centering
    \includegraphics[width=\columnwidth]{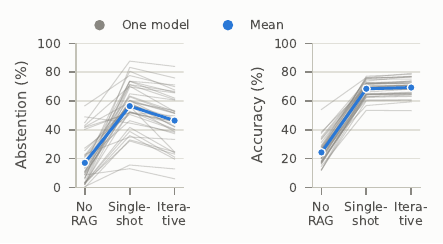}
    \caption{Correct Abstention Rate (left) and Answer Accuracy (right) without retrieval, with single-shot RAG and with iterative RAG: one gray line per configuration, mean in blue. Single-shot RAG raises CAR significantly for \DCMainRagUp{} of \DCMainRagPairs{} configurations and lowers it for \DCMainRagDown; iterative RAG lowers it for \DCMainItDown{} and raises it for \DCMainItUp{} ($q<0.05$). Accuracy rises with retrieval and barely changes with further rounds.}
    \label{fig:retrieval}
\end{figure}

\begin{figure}[!htbp]
    \centering
    \includegraphics[width=\columnwidth]{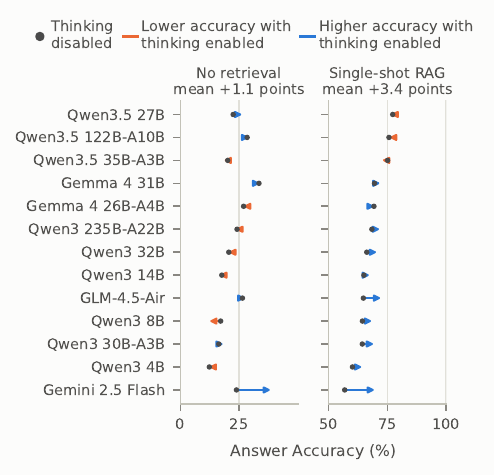}
    \caption{Answer Accuracy with thinking disabled (gray dot) and enabled (arrowhead) on the same weights, without retrieval (left) and with single-shot RAG (right), for the models of \Cref{fig:thinking}. Orange: lower with thinking enabled; blue: higher. Each panel spans 50\% points.}
    \label{fig:thinking_acc}
\end{figure}

\begin{figure}[!htbp]
    \centering
    \includegraphics[width=\columnwidth]{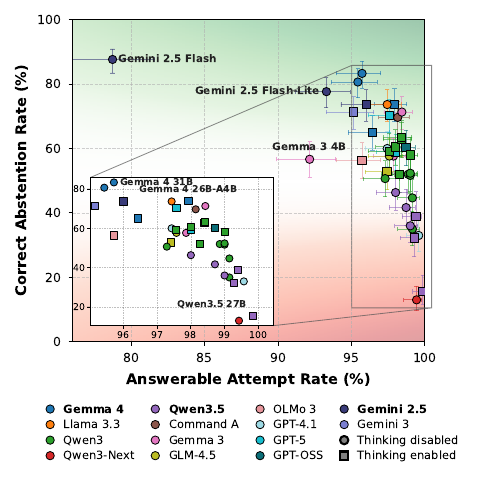}
    \caption{Answerable Attempt Rate against Correct Abstention Rate for every configuration, single-shot RAG. Colors mark model families; squares are runs with thinking enabled, circles runs with thinking disabled or from instruct-only models. Bars are 95\% Wilson intervals. The inset magnifies the crowded high-coverage region (without error bars); labels mark each family's highest Correct Abstention Rate and every configuration that no other configuration exceeds on both rates. Attempt rates span \DCAttemptMin--\DCAttemptMax\%; CAR spans \DCAbstainMin--\DCAbstainMax\%. The best performance is the top-right corner.}
    \label{fig:scatter}
\end{figure}

\begin{figure}[!htbp]
    \centering
    \includegraphics[width=\columnwidth]{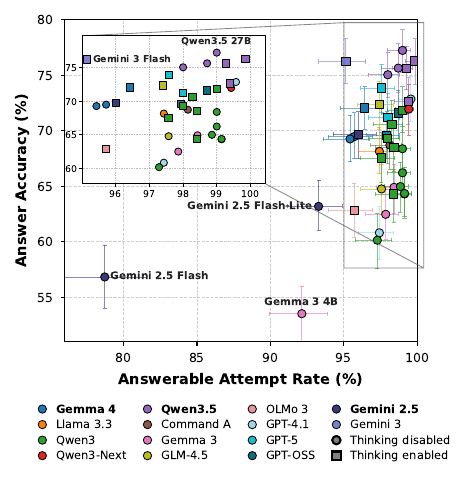}
    \caption{Answerable Attempt Rate against Answer Accuracy (mean of the three accuracy judges), single-shot RAG. Markers and inset as in \Cref{fig:scatter}. Bars are 95\% intervals: Wilson for the attempt rate, bootstrap over questions for accuracy.}
    \label{fig:acc_scatter}
\end{figure}

\begin{figure}[!htbp]
    \centering
    \includegraphics[width=\columnwidth]{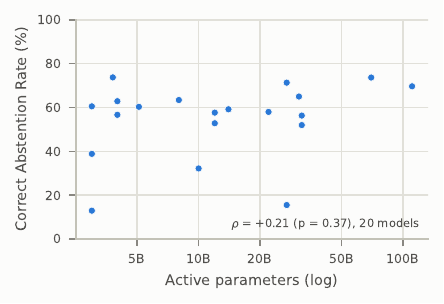}
    \caption{Correct Abstention Rate versus active parameters across the \DFNTotal{} open-weight models, with single-shot RAG; one point per model as in \Cref{fig:size}. The Spearman correlation is not significant at the 5\% level.}
    \label{fig:size_active}
\end{figure}

\begin{figure}[!htbp]
    \centering
    \includegraphics[width=\columnwidth]{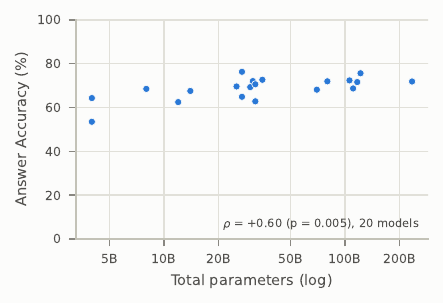}
    \caption{Answer Accuracy versus total parameters across the \DFNTotal{} open-weight models, with single-shot RAG; one point per model as in \Cref{fig:size}. Unlike Correct Abstention Rate, Answer Accuracy rises with size (Spearman $\rho=\DFRhoAccTotal$, $p=\DFPAccTotal$).}
    \label{fig:size_accuracy}
\end{figure}

\paragraph{Views of the main findings per model.} \Cref{fig:retrieval} shows Correct Abstention Rate and Answer Accuracy under all three retrieval conditions for every configuration. \Cref{fig:thinking_acc} shows the accuracy changes from enabling thinking, and \Cref{fig:scatter,fig:acc_scatter} place individual configurations by their Answerable Attempt Rate, Correct Abstention Rate and Answer Accuracy. \Cref{fig:size_active,fig:size_accuracy} repeat the size comparison of \Cref{fig:size} with active parameters and with Answer Accuracy.

\begin{figure*}[!tp]
    \centering
    \includegraphics[width=\textwidth,height=0.85\textheight,keepaspectratio]{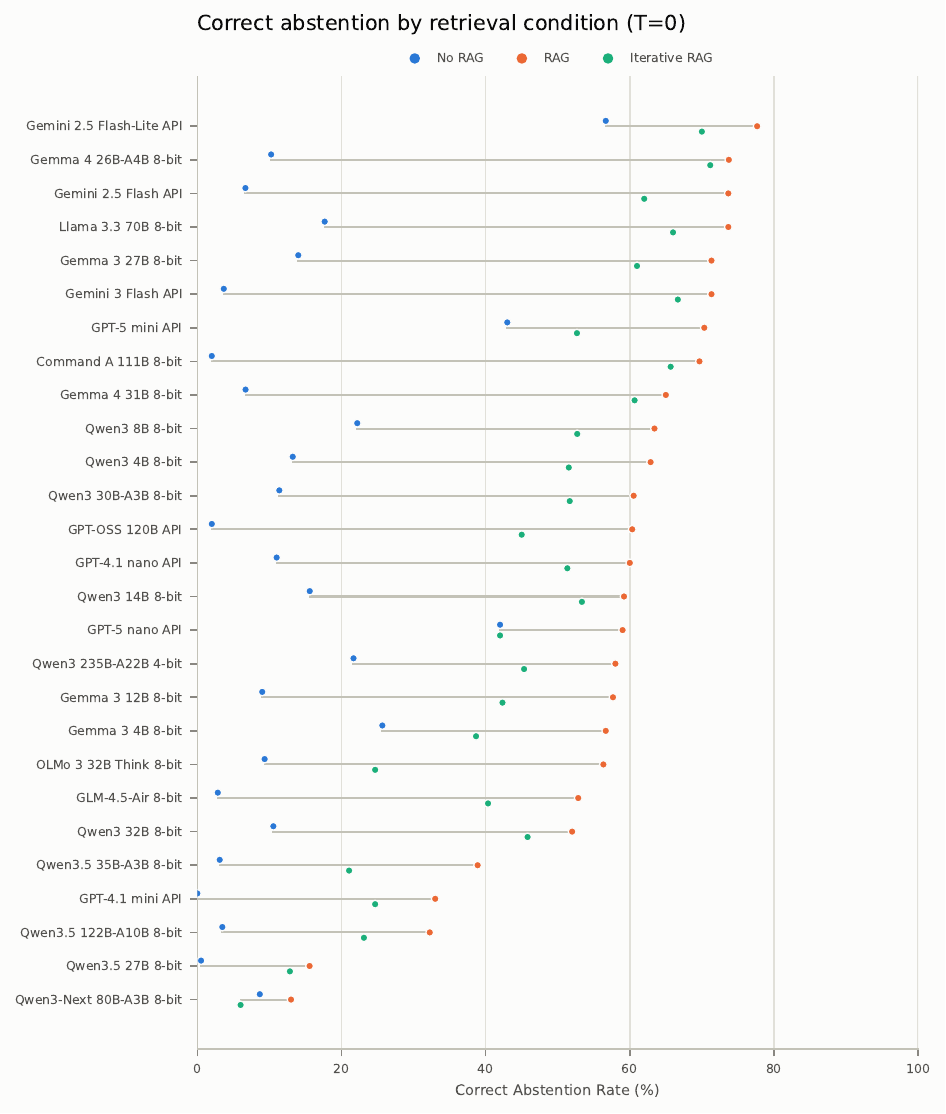}
    \caption{Correct Abstention Rate per model without retrieval (No RAG), with single-shot RAG and with iterative RAG, thinking enabled or instruct-only. Across all configurations, single-shot RAG raises CAR for \DCRagUp{} of \DCRagPairs{} and lowers it for \DCRagDown; iterative RAG lowers it for \DCItDown{} of \DCItPairs{} relative to single-shot RAG ($q<0.05$), by \DCItAbstDelta{}\% points on average.}
    \label{fig:RAG02}
\end{figure*}

\begin{figure*}[!tp]
    \centering
    \includegraphics[width=\textwidth,height=0.85\textheight,keepaspectratio]{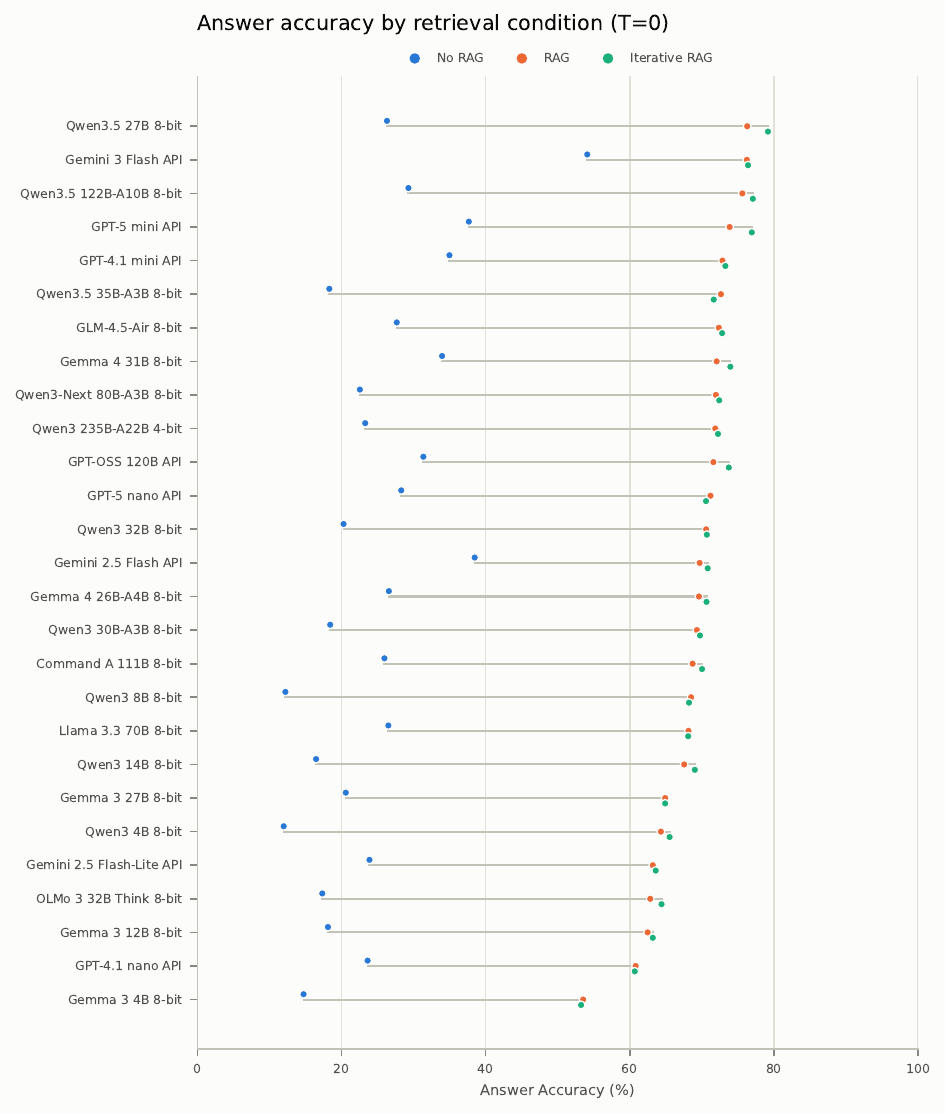}
    \caption{Answer Accuracy per model under the three retrieval conditions, thinking enabled or instruct-only. Mean accuracy is \DCAccNoRag\% without retrieval, \DCAccRag\% with single-shot RAG and \DCAccIter\% with iterative RAG.}
    \label{fig:RAG02acc}
\end{figure*}

\begin{figure}[!htbp]
    \centering
    \includegraphics[width=\columnwidth]{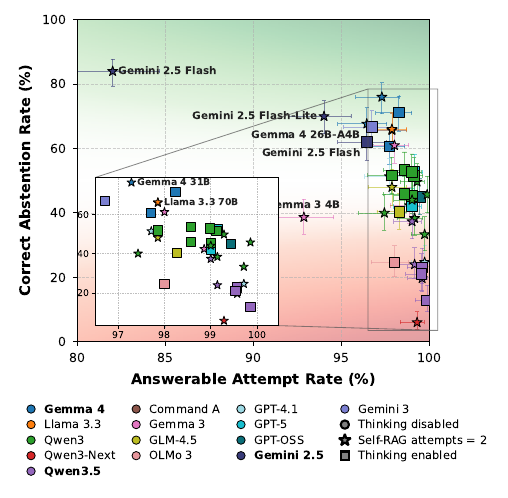}
    \caption{As \Cref{fig:scatter}, with iterative RAG (up to two further retrieval rounds).}
    \label{fig:scatter_iterative}
\end{figure}

\paragraph{Retrieval conditions.} Single-shot RAG raises CAR for most configurations, and iterative RAG lowers it again (\Cref{fig:RAG02}); accuracy rises sharply with retrieval and changes little with further rounds (\Cref{fig:RAG02acc}). Iterative RAG lowers Correct Abstention Rate at an almost unchanged Answerable Attempt Rate (\Cref{fig:scatter_iterative}, against \Cref{fig:scatter}). Every model and thinking setting under every condition is in \Cref{tab:results_rag0,tab:results_rag2}.

\begin{figure}[!htbp]
    \centering
    \includegraphics[width=\columnwidth]{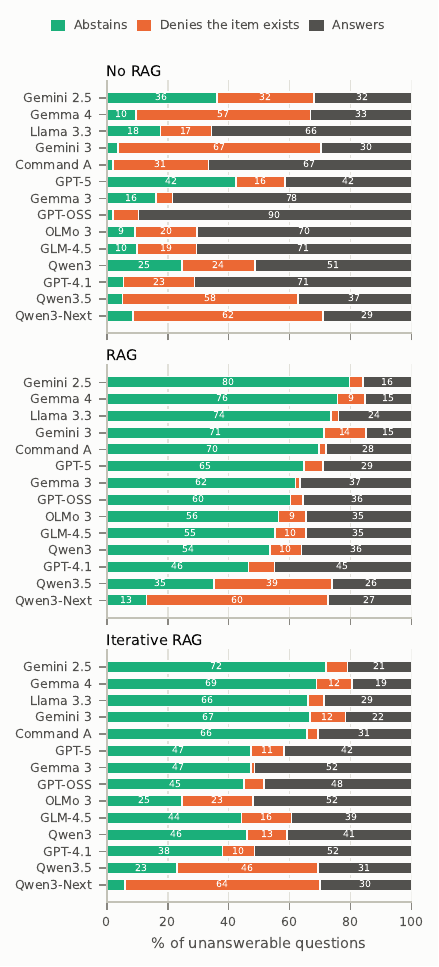}
    \caption{What each model family does with unanswerable questions, by retrieval condition: abstention, premise denial or attempted answer. Family means over configurations.}
    \label{fig:outcomes}
\end{figure}

\begin{figure}[!htbp]
    \centering
    \includegraphics[width=\columnwidth]{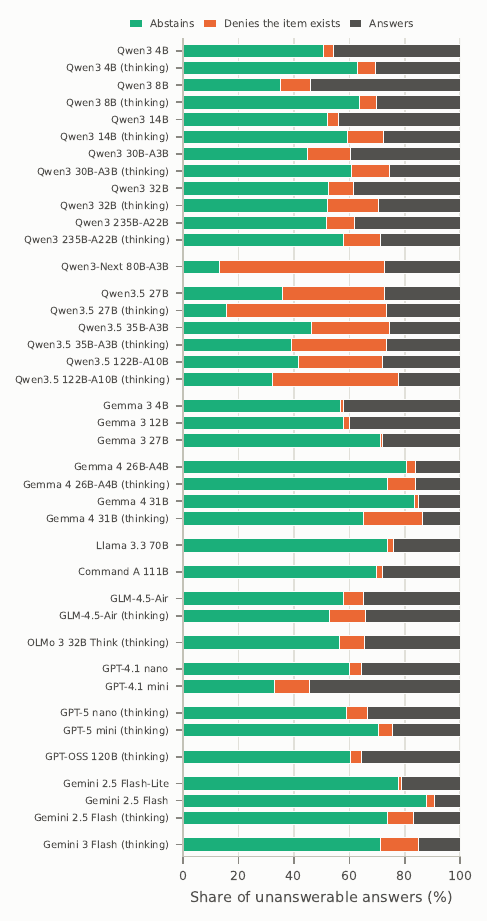}
    \caption{What each model does with the unanswerable questions under single-shot RAG: abstention, premise denial or attempted answer. Families together; within a family, smallest model first.}
    \label{fig:outcomes_rag}
\end{figure}

\paragraph{How models decline.} A model that declines either abstains or responds with premise denial, and the mix differs by family (\Cref{fig:outcomes}). Premise denials make up \DCDenyNoRag\% of responses to unanswerable questions without retrieval, \DCDenyRag\% with single-shot RAG and \DCDenyIter\% with iterative RAG. Without retrieval, \DCFamDenyMoreNoRag{} of the \DCFamNNoRag{} families give premise denials more often than abstentions. With single-shot RAG, only \DCFamDenyMoreRag{} do, Qwen3.5 and Qwen3-Next (each model: \Cref{fig:outcomes_rag}), and the other families mostly say they do not know. Counting premise denials as abstentions would therefore change some families' rates far more than others' (\Cref{sec:appendix-judges}).

\begin{figure}[!htbp]
    \centering
    \includegraphics[width=\columnwidth]{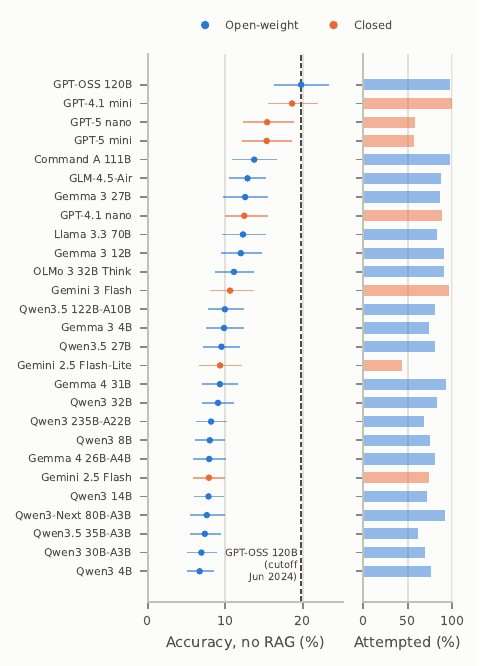}
    \caption{Leakage audit per model: Answer Accuracy on the unanswerable questions without retrieval (mean of the three accuracy judges against the oracle model's reference answers from $C_\text{post}$, an abstention scoring 0), with 95\% bootstrap intervals over questions, next to each model's attempt rate. The dashed line is \texttt{GPT-OSS 120B}, whose published knowledge cut-off (June 2024) predates every post-cut-off page.}
    \label{fig:leakage_model}
\end{figure}

\paragraph{Leakage.} Without retrieval, \texttt{GPT-OSS 120B}, whose published knowledge cut-off predates every post-cut-off page, answers the unanswerable questions most accurately (\DCLeakGPTOSS\%), followed by GPT-4.1 (\DCLeakGPTFourOne\%). These models attempt most of the questions, and every other model (including ones without a published knowledge cut-off) is less accurate. Accuracy here is Answer Accuracy on the unanswerable questions: the mean of the three accuracy judges' scores against the reference answers that the oracle model writes from the $C_\text{post}$ pages, with an abstention or a premise denial scoring 0. \Cref{fig:leakage_model} and \Cref{tab:leakage} show it for each model and family. These answers are best explained as correct guesses from older, general mechanics rather than knowledge of the new pages. No model without a published knowledge cut-off exceeds this reference, so the audit finds no evidence that these models learned the $C_\text{post}$ pages.

\begin{table*}[t]
\centering
\scriptsize
\begin{tabular}{lllrrrr}
\toprule
Family & Runs & Attempted \% & Correct \% & CI low & CI high & Accuracy when attempted \% \\
\midrule
GPT-OSS & 1 & 97.7 & 19.8 & 16.6 & 23.0 & 20.3 \\
GPT-4.1 & 2 & 94.5 & 15.5 & 13.0 & 17.9 & 16.4 \\
GPT-5 & 2 & 57.5 & 15.4 & 12.7 & 18.3 & 26.7 \\
Command A & 1 & 98.0 & 13.7 & 11.0 & 16.5 & 14.0 \\
GLM-4.5 & 2 & 87.2 & 12.9 & 10.6 & 15.1 & 14.8 \\
Llama 3.3 & 1 & 82.3 & 12.3 & 9.6 & 15.2 & 14.9 \\
Gemma 3 & 3 & 83.8 & 11.5 & 9.5 & 13.4 & 13.7 \\
OLMo 3 & 1 & 90.7 & 11.1 & 8.7 & 13.7 & 12.3 \\
Gemini 3 & 1 & 96.3 & 10.6 & 8.0 & 13.2 & 11.0 \\
Qwen3.5 & 6 & 74.5 & 8.9 & 7.2 & 10.9 & 12.0 \\
Gemma 4 & 4 & 86.9 & 8.6 & 6.6 & 10.6 & 9.9 \\
Gemini 2.5 & 3 & 63.9 & 8.4 & 6.7 & 10.2 & 13.1 \\
Qwen3 & 12 & 73.9 & 7.8 & 6.4 & 9.2 & 10.5 \\
Qwen3-Next & 1 & 91.3 & 7.6 & 5.5 & 9.7 & 8.3 \\
\bottomrule
\end{tabular}
\caption{Leakage audit by model family: unanswerable questions answered correctly without retrieval, with 95\% bootstrap intervals over questions.}
\label{tab:leakage}
\end{table*}

\begin{figure*}[t]
    \centering
    \includegraphics[width=0.9\textwidth]{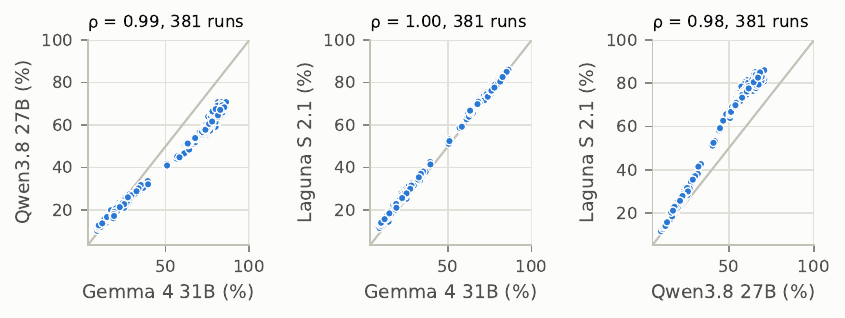}
    \caption{Run-level Answer Accuracy under each pair of accuracy judges (Spearman $\rho$ above each panel). The judges rank runs almost identically, but their mean Answer Accuracy differs by up to \DCJudgeSpread{}\% points.}
    \label{fig:judges}
\end{figure*}

\paragraph{Accuracy judges.} The three accuracy judges rank runs almost identically but differ in leniency (how high a judge scores the same answers): their mean Answer Accuracy differs by up to \DCJudgeSpread{}\% points (\Cref{fig:judges}; Fleiss' $\kappa = \DCFleissKappa$, \Cref{tab:judges}), so accuracy is reported as their mean and orderings matter more than absolute values. \Cref{sec:res-judge-selfpref} tests whether a judge favors its own model family.

\begin{table*}[t]
\centering
\scriptsize
\begin{tabular}{lrrrr}
\toprule
Judges & Rows & Exact agreement \% & Weighted $\kappa$ & Mean diff (pts) \\
\midrule
Gemma 4 31B (8-bit) vs Qwen3.8 27B (8-bit) & 252,820 & 68.5 & 0.730 & +10.9 \\
Gemma 4 31B (8-bit) vs Laguna S 2.1 (8-bit) & 252,820 & 85.2 & 0.888 & -1.9 \\
Qwen3.8 27B (8-bit) vs Laguna S 2.1 (8-bit) & 252,820 & 69.0 & 0.728 & -12.8 \\
\bottomrule
\end{tabular}
\caption{Agreement between the three accuracy judges on the same answers (quadratic-weighted Cohen's $\kappa$; Fleiss' $\kappa$ across all three is \DCFleissKappa).}
\label{tab:judges}
\end{table*}

\subsection{Do Accuracy Judges Have a Self-Preference?}
\label{sec:res-judge-selfpref}
Two of the three accuracy judges come from model families we also evaluate: \texttt{Gemma 4 31B} (Gemma) and \texttt{Qwen3.8 27B} (Qwen). A judge with a self-preference would score answers from its own family higher than the other judges do. For each of these two judges, we take the difference between its score and the mean score of the other two judges on the same answer. Its self-preference is the mean difference on answers from models of its own family minus the mean difference on answers from all other models, with a 95\% bootstrap interval over questions.

Both judges show a small self-preference: \DCSelfPrefGemmaFourThreeOneBEightbit\% points for \texttt{Gemma 4 31B} (\DCSelfPrefGemmaFourThreeOneBEightbitLo{} to \DCSelfPrefGemmaFourThreeOneBEightbitHi) and \DCSelfPrefQwenThreeEightTwoSevenBEightbit\% points for \texttt{Qwen3.8 27B} (\DCSelfPrefQwenThreeEightTwoSevenBEightbitLo{} to \DCSelfPrefQwenThreeEightTwoSevenBEightbitHi). Both are small next to the judges' differences in leniency, a judge's tendency to score every answer higher or lower than the other judges do, whatever model wrote it. Against the mean of the three judges, \texttt{Gemma 4 31B} scores \DCSeverityGemmaFourThreeOneBEightbit, \texttt{Laguna S 2.1} \DCSeverityLagunaSTwoOneEightbit{} and \texttt{Qwen3.8 27B} \DCSeverityQwenThreeEightTwoSevenBEightbit\% points. Because Answer Accuracy averages the three judges, a self-preference enters it at about a third of its size.

As a further check, we recompute each configuration's Answer Accuracy using only the accuracy judges from other families. The ranking of the \DFNAccAbst{} configurations with single-shot RAG then has a Spearman correlation of \DCRhoLeaveFamilyOutRag{} with the original ranking. The change comes mostly from leniency rather than self-preference. Dropping \texttt{Qwen3.8 27B}, the least lenient judge, changes the accuracy of Qwen3, Qwen3-Next and Qwen3.5 models by \DCLooShiftQwenThree, \DCLooShiftQwenThreeNext{} and \DCLooShiftQwenThreeFive\% points on average, and dropping \texttt{Gemma 4 31B}, which scores above the mean, changes Gemma~3 and Gemma~4 models by \DCLooShiftGemmaThree{} and \DCLooShiftGemmaFour\% points.

\subsection{Ablation: Increasing Retrieval Depth Lowers Correct Abstention Rate}
\label{sec:res-depth}
To compare a larger single retrieval with iterative RAG, we evaluate the \DFDepthModels{} models with single-shot runs at both $k{=}5$ and $k{=}10$ (\Cref{fig:depth}). We disable thinking where switchable and use the other models' defaults. On this subset, mean Correct Abstention Rate is \DFDepthNoRag\% without retrieval, \DFDepthKFive\% at $k{=}5$, \DFDepthKTen\% at $k{=}10$ and \DFDepthIter\% with iterative RAG. Iterative RAG yields a lower Correct Abstention Rate than single-shot RAG at $k{=}5$ for all \DFDepthIterLower{} models. Doubling $k$ also lowers the mean, so the results do not support attributing the iterative decrease solely to follow-up query selection: chunk count and context length change together, and iterative RAG additionally changes the queries and introduces self-reflection.

\begin{figure}[!htbp]
    \centering
    \includegraphics[width=\columnwidth]{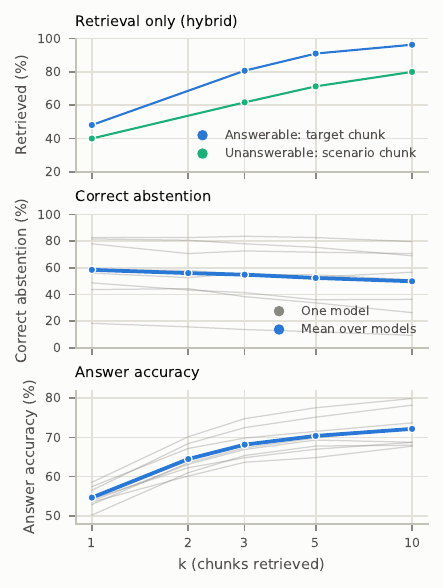}
    \caption{Choice of $k$ (hybrid retrieval, single-shot RAG). Top: retrieval alone, the share of answerable questions whose target chunk is retrieved and of unanswerable questions for which a chunk the scenario was written from is retrieved. Middle and bottom: Correct Abstention Rate and Answer Accuracy for \DCKModels{} models run at every $k$ (gray; thinking disabled where switchable) and their mean (blue). From $k{=}5$ to $k{=}10$, CAR falls from \DCKAbstFive\% to \DCKAbstOneZero\% while accuracy moves from \DCKAccFive\% to \DCKAccOneZero\%.}
    \label{fig:ksweep}
\end{figure}

\begin{table}[!htbp]
\setlength{\tabcolsep}{4pt}
\centering
\scriptsize
\begin{tabular}{lrrrr}
\toprule
Model & k & CAR \% & Accuracy \% & Bal. F1 \\
\midrule
Command A 111B & 1 & \textbf{78.0} & 53.8 & 82.2 \\
Command A 111B & 2 & 70.7 & 63.1 & 84.0 \\
Command A 111B & 3 & 72.7 & 66.9 & 86.3 \\
Command A 111B & 5 & 71.7 & \textbf{69.3} & \textbf{86.6} \\
Command A 111B & 10 & 70.7 & 68.8 & 86.1 \\
GLM-4.5-Air & 1 & \textbf{58.7} & 53.5 & 78.1 \\
GLM-4.5-Air & 2 & 58.0 & 60.1 & 80.3 \\
GLM-4.5-Air & 3 & 55.0 & 63.7 & 80.2 \\
GLM-4.5-Air & 5 & 54.7 & 64.9 & \textbf{80.5} \\
GLM-4.5-Air & 10 & 51.0 & \textbf{67.8} & 79.4 \\
Gemma 4 31B & 1 & 82.7 & 53.0 & 79.8 \\
Gemma 4 31B & 2 & 82.7 & 63.5 & 86.7 \\
Gemma 4 31B & 3 & \textbf{83.7} & 67.3 & 88.8 \\
Gemma 4 31B & 5 & 82.7 & 69.9 & \textbf{89.9} \\
Gemma 4 31B & 10 & 79.7 & \textbf{72.3} & 89.8 \\
Llama 3.3 70B & 1 & \textbf{81.7} & 50.2 & 79.6 \\
Llama 3.3 70B & 2 & 80.7 & 61.0 & 86.1 \\
Llama 3.3 70B & 3 & 78.0 & 65.5 & 87.4 \\
Llama 3.3 70B & 5 & 75.3 & 67.9 & \textbf{87.6} \\
Llama 3.3 70B & 10 & 69.0 & \textbf{68.0} & 85.7 \\
Qwen3 32B & 1 & 56.0 & 54.3 & 77.9 \\
Qwen3 32B & 2 & 52.7 & 62.3 & 79.7 \\
Qwen3 32B & 3 & 56.0 & 64.8 & 81.1 \\
Qwen3 32B & 5 & 53.0 & 67.0 & 80.5 \\
Qwen3 32B & 10 & \textbf{56.7} & \textbf{68.7} & \textbf{81.6} \\
Qwen3-Next 80B-A3B & 1 & \textbf{18.3} & 57.5 & 68.5 \\
Qwen3-Next 80B-A3B & 2 & 15.7 & 67.2 & \textbf{69.7} \\
Qwen3-Next 80B-A3B & 3 & 13.7 & 69.8 & 69.2 \\
Qwen3-Next 80B-A3B & 5 & 12.0 & 71.5 & 69.1 \\
Qwen3-Next 80B-A3B & 10 & 9.3 & \textbf{73.7} & 68.5 \\
Qwen3.5 122B-A10B & 1 & \textbf{48.7} & 56.5 & 71.9 \\
Qwen3.5 122B-A10B & 2 & 43.3 & 68.4 & 75.8 \\
Qwen3.5 122B-A10B & 3 & 41.3 & 72.5 & \textbf{76.1} \\
Qwen3.5 122B-A10B & 5 & 36.0 & 75.2 & 74.9 \\
Qwen3.5 122B-A10B & 10 & 36.3 & \textbf{78.2} & 75.6 \\
Qwen3.5 27B & 1 & 43.7 & 58.5 & 72.2 \\
Qwen3.5 27B & 2 & \textbf{44.3} & 70.2 & \textbf{76.4} \\
Qwen3.5 27B & 3 & 38.3 & 74.8 & 75.3 \\
Qwen3.5 27B & 5 & 33.7 & 77.6 & 74.6 \\
Qwen3.5 27B & 10 & 26.3 & \textbf{79.9} & 72.9 \\
\bottomrule
\end{tabular}
\caption{$k$ sweep per model (hybrid retrieval, single-shot RAG; the models of \Cref{fig:ksweep}). \emph{Bal.\ F1}: balanced F1 (\Cref{sec:appendix-class-balance}). Bold: the best $k$ for each model in each column. The $k{=}5$ rows are the hybrid runs of \Cref{tab:ablation}, not the runs in \Cref{tab:results_rag0}.}
\label{tab:ksweep}
\end{table}

A separate sweep over \DCKModels{} models covers $k\in\{1,2,3,5,10\}$ (\Cref{fig:ksweep,tab:ksweep}). In that subset, moving from five to ten chunks lowers mean Correct Abstention Rate from \DCKAbstFive\% to \DCKAbstOneZero\%, while mean Answer Accuracy rises from \DCKAccFive\% to \DCKAccOneZero\%. These means describe the smaller sweep, not the \DFDepthModels{}-model comparison above.

\subsection{Ablation: Query Rewriting and Reranking Lower Correct Abstention Rate}
\label{sec:res-pipeline}
We compare six pipelines under single-shot RAG with five retrieved chunks, holding the evaluated models and their thinking settings fixed. Appendix~\ref{sec:appendix-retrieval} specifies the retrievers, query rewrites and reranker. For the \DFPipeModels{} models in \Cref{fig:pipeline}, mean Correct Abstention Rate is \DFPipeBm\% with BM25, \DFPipeDense\% with dense retrieval, \DFPipeHybrid\% with the default hybrid retriever, \DFPipeRerank\% with hybrid plus reranking, \DFPipeRewrite\% with hybrid plus query rewriting and \DFPipeFull\% with both additions. The combined pipeline yields a lower Correct Abstention Rate than the default hybrid retriever for all \DFPipeFullBelow{} models. These results concern the tested implementations; pipeline complexity does not establish that one retriever supplies better evidence.

\begin{figure}[!htbp]
    \centering
    \includegraphics[width=\columnwidth]{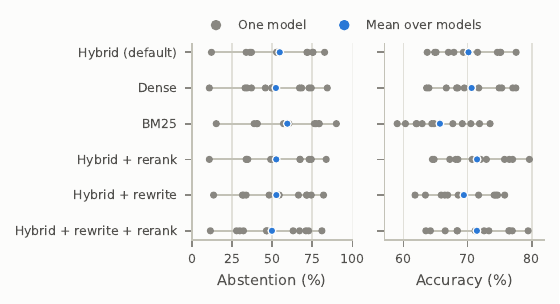}
    \caption{Retrieval ablation: Correct Abstention Rate (left) and Answer Accuracy (right) for \DFPipeModels{} models under six retrieval pipelines (single-shot RAG, $k{=}5$; thinking disabled where switchable, \texttt{OLMo 3 32B Think} and closed models at their defaults). Gray: individual models; blue: their mean. Mean Correct Abstention Rate is \DCAblHybridAbst\% with the default hybrid retriever, \DCAblRerankAbst\% with reranking and \DCAblFullAbst\% with query rewriting and reranking; \DCAblSig{} of \DCAblTests{} model $\times$ pipeline comparisons with hybrid are significant.}
    \label{fig:ablation}
\end{figure}

\begin{table*}[!tp]
\centering
\scriptsize
\begin{tabular}{lllrrr}
\toprule
Model & Thinking & Retrieval & CAR \% & Accuracy \% & Bal. F1 \\
\midrule
Command A 111B & no & Hybrid (default) & 71.7 & 69.3 & 86.6 \\
Command A 111B & no & Dense & 68.3 & 69.5 & 85.7 \\
Command A 111B & no & BM25 & \textbf{77.3} & 64.8 & \textbf{87.0} \\
Command A 111B & no & Hybrid + rerank & 67.3 & 70.7 & 85.2 \\
Command A 111B & no & Hybrid + rewrite & 71.7 & 68.5 & 86.2 \\
Command A 111B & no & Hybrid + rewrite + rerank & 67.0 & \textbf{71.2} & 84.9 \\
GLM-4.5-Air & no & Hybrid (default) & 54.7 & 64.9 & 80.5 \\
GLM-4.5-Air & no & Dense & 45.7 & 66.7 & 78.0 \\
GLM-4.5-Air & no & BM25 & \textbf{60.3} & 62.0 & \textbf{81.3} \\
GLM-4.5-Air & no & Hybrid + rerank & 51.3 & \textbf{67.2} & 80.0 \\
GLM-4.5-Air & no & Hybrid + rewrite & 52.0 & 65.9 & 79.9 \\
GLM-4.5-Air & no & Hybrid + rewrite + rerank & 46.7 & 66.5 & 78.4 \\
GPT-5 mini & yes & Hybrid (default) & 72.0 & 74.6 & \textbf{86.3} \\
GPT-5 mini & yes & Dense & 67.0 & 74.9 & 84.6 \\
GPT-5 mini & yes & BM25 & \textbf{76.3} & 69.2 & 85.1 \\
GPT-5 mini & yes & Hybrid + rerank & 67.3 & 75.8 & 85.0 \\
GPT-5 mini & yes & Hybrid + rewrite & 66.3 & 74.7 & 84.5 \\
GPT-5 mini & yes & Hybrid + rewrite + rerank & 63.0 & \textbf{76.4} & 83.5 \\
Gemini 3 Flash & yes & Hybrid (default) & 75.3 & 75.0 & 86.5 \\
Gemini 3 Flash & yes & Dense & 74.3 & \textbf{77.0} & 87.2 \\
Gemini 3 Flash & yes & BM25 & \textbf{79.3} & 70.5 & 85.7 \\
Gemini 3 Flash & yes & Hybrid + rerank & 74.3 & \textbf{77.0} & \textbf{87.3} \\
Gemini 3 Flash & yes & Hybrid + rewrite & 74.3 & 74.1 & 86.4 \\
Gemini 3 Flash & yes & Hybrid + rewrite + rerank & 72.7 & 76.5 & 86.0 \\
Gemma 4 31B & no & Hybrid (default) & 82.7 & 69.9 & 89.9 \\
Gemma 4 31B & no & Dense & 84.3 & 70.5 & 91.1 \\
Gemma 4 31B & no & BM25 & \textbf{90.0} & 64.4 & 89.7 \\
Gemma 4 31B & no & Hybrid + rerank & 83.7 & 72.0 & \textbf{91.4} \\
Gemma 4 31B & no & Hybrid + rewrite & 82.0 & 69.3 & 89.6 \\
Gemma 4 31B & no & Hybrid + rewrite + rerank & 81.0 & \textbf{72.6} & 90.3 \\
Llama 3.3 70B & no & Hybrid (default) & 75.3 & 67.9 & \textbf{87.6} \\
Llama 3.3 70B & no & Dense & 73.0 & \textbf{68.5} & 86.8 \\
Llama 3.3 70B & no & BM25 & \textbf{79.0} & 62.0 & 86.4 \\
Llama 3.3 70B & no & Hybrid + rerank & 73.0 & 68.1 & 87.1 \\
Llama 3.3 70B & no & Hybrid + rewrite & 71.3 & 66.9 & 85.7 \\
Llama 3.3 70B & no & Hybrid + rewrite + rerank & 71.0 & 68.4 & 86.3 \\
OLMo 3 32B Think & yes & Hybrid (default) & 52.7 & 63.7 & 79.0 \\
OLMo 3 32B Think & yes & Dense & 49.7 & 63.6 & 78.8 \\
OLMo 3 32B Think & yes & BM25 & \textbf{58.7} & 59.0 & \textbf{79.4} \\
OLMo 3 32B Think & yes & Hybrid + rerank & 50.7 & \textbf{64.5} & 78.8 \\
OLMo 3 32B Think & yes & Hybrid + rewrite & 48.0 & 61.8 & 78.0 \\
OLMo 3 32B Think & yes & Hybrid + rewrite + rerank & 49.0 & 63.5 & 78.2 \\
Qwen3 32B & no & Hybrid (default) & 53.0 & 67.0 & 80.5 \\
Qwen3 32B & no & Dense & 49.7 & 68.3 & 79.6 \\
Qwen3 32B & no & BM25 & \textbf{57.0} & 62.9 & \textbf{80.8} \\
Qwen3 32B & no & Hybrid + rerank & 49.0 & \textbf{68.5} & 79.3 \\
Qwen3 32B & no & Hybrid + rewrite & 54.3 & 66.4 & 80.6 \\
Qwen3 32B & no & Hybrid + rewrite + rerank & 46.3 & 68.4 & 78.4 \\
Qwen3 8B & no & Hybrid (default) & 37.0 & \textbf{65.1} & \textbf{75.9} \\
Qwen3 8B & no & Dense & 37.0 & 64.0 & 75.7 \\
Qwen3 8B & no & BM25 & \textbf{38.7} & 60.3 & 75.4 \\
Qwen3 8B & no & Hybrid + rerank & 33.7 & 64.8 & 74.8 \\
Qwen3 8B & no & Hybrid + rewrite & 33.7 & 63.4 & 74.8 \\
Qwen3 8B & no & Hybrid + rewrite + rerank & 32.0 & 64.2 & 74.4 \\
Qwen3-Next 80B-A3B & no & Hybrid (default) & 12.0 & 71.5 & 69.1 \\
Qwen3-Next 80B-A3B & no & Dense & 10.7 & 71.7 & 68.9 \\
Qwen3-Next 80B-A3B & no & BM25 & \textbf{15.0} & 67.7 & \textbf{69.7} \\
Qwen3-Next 80B-A3B & no & Hybrid + rerank & 10.7 & 72.9 & 68.9 \\
Qwen3-Next 80B-A3B & no & Hybrid + rewrite & 13.3 & 71.7 & 69.3 \\
Qwen3-Next 80B-A3B & no & Hybrid + rewrite + rerank & 11.3 & \textbf{73.3} & 68.9 \\
Qwen3.5 122B-A10B & no & Hybrid (default) & 36.0 & 75.2 & 74.9 \\
Qwen3.5 122B-A10B & no & Dense & 33.0 & 75.3 & 74.2 \\
Qwen3.5 122B-A10B & no & BM25 & \textbf{40.7} & 71.9 & \textbf{75.3} \\
Qwen3.5 122B-A10B & no & Hybrid + rerank & 34.7 & 76.5 & 75.2 \\
Qwen3.5 122B-A10B & no & Hybrid + rewrite & 31.3 & 74.4 & 73.8 \\
Qwen3.5 122B-A10B & no & Hybrid + rewrite + rerank & 29.7 & \textbf{77.0} & 73.5 \\
Qwen3.5 27B & no & Hybrid (default) & 33.7 & 77.6 & 74.6 \\
Qwen3.5 27B & no & Dense & 34.3 & 77.5 & 75.1 \\
Qwen3.5 27B & no & BM25 & \textbf{40.3} & 73.5 & \textbf{75.4} \\
Qwen3.5 27B & no & Hybrid + rerank & 34.0 & \textbf{79.6} & 75.0 \\
Qwen3.5 27B & no & Hybrid + rewrite & 32.0 & 75.8 & 74.2 \\
Qwen3.5 27B & no & Hybrid + rewrite + rerank & 27.7 & 79.5 & 73.2 \\
\bottomrule
\end{tabular}
\caption{Retrieval ablation per model (single-shot RAG, $k{=}5$). \emph{Thinking}: whether thinking is enabled. \emph{Bal.\ F1}: balanced F1 (\Cref{sec:appendix-class-balance}). Bold: the best pipeline for each model in each column. The \emph{Hybrid (default)} rows are a second run of the configurations in \Cref{tab:results_rag0}, with a different set of responses.}
\label{tab:ablation}
\end{table*}

\Cref{tab:ablation} gives every tested model and \Cref{fig:ablation} plots both Correct Abstention Rate and Answer Accuracy. Over the \DFPipeModels{} models in that analysis, the mean Correct Abstention Rate falls from \DCAblHybridAbst\% with hybrid retrieval to \DCAblFullAbst\% with both rewriting and reranking, while mean Answer Accuracy changes from \DCAblHybridAcc\% to \DCAblFullAcc\%. Only \DCAblSig{} of the \DCAblTests{} paired comparisons of alternative pipelines against hybrid retrieval are significant. The all-model direction of the combined-pipeline result above should therefore not be read as a significant decrease for every individual model.

\subsection{Ablation: Scenarios Make Abstention Harder}
\label{sec:res-scenario}
\begin{figure}[!htbp]
    \centering
    \includegraphics[width=\columnwidth]{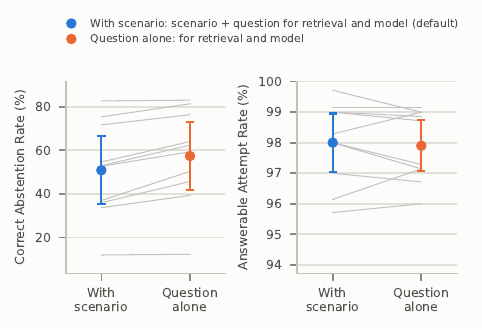}
    \caption{The scenario makes abstention harder. Correct Abstention Rate and Answerable Attempt Rate for \DFScenModelsStrict{} models (single-shot RAG, $k{=}5$) with the scenario (blue: scenario and question for retrieval and the model; the default) and without it (orange: the question alone for both). Points: mean with a 95\% confidence interval over models; gray lines: one model. Each panel has its own scale.}
    \label{fig:scenario}
\end{figure}

\begin{figure}[!htbp]
    \centering
    \includegraphics[width=\columnwidth]{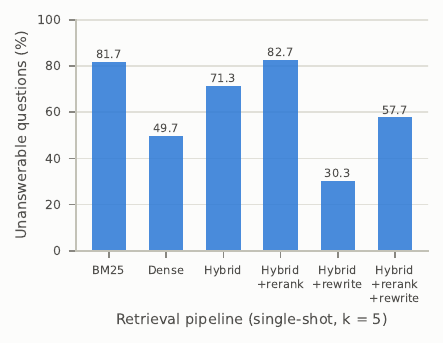}
    \caption{Share of unanswerable questions whose five retrieved chunks include a $C_\text{pre}$ passage the scenario was written from, by retrieval pipeline (retrieval only).}
    \label{fig:scenario_passages}
\end{figure}

Each question opens with a player scenario written from $C_\text{pre}$ pages (\Cref{sec:scenario-construction}). To test whether it matters, we reran \DFScenModelsStrict{} models without it: the question alone serves as both the retrieval query and the model's input. Without the scenario, Correct Abstention Rate rises from \DFScenDefaultStrict\% to \DFScenQonlyStrict\%, for all \DFScenQonlyHigherStrict{} models, while Answerable Attempt Rate is almost unchanged (\Cref{fig:scenario}). A benchmark of bare questions would therefore overstate how often models abstain. With hybrid retrieval, \DCPremiseHitFive\% of unanswerable questions also retrieve a passage the scenario was written from (\Cref{fig:scenario_passages}).

\subsection{Thinking: Family Differences and Response Categories}
\label{sec:res-thinking-details}
The paired comparison in \Cref{sec:res-thinking} uses the \DFThinkModels{} models with switchable thinking, with the same weights and questions in each pair. Without retrieval, enabling thinking lowers mean Correct Abstention Rate by \DFThinkNoRagMeanMag{}\% points (Wilcoxon signed-rank test over per-model differences, $p=\DFThinkNoRagP$). With single-shot RAG, the mean change is $\DFThinkRagMean$\% points ($p=\DFThinkRagP$, Wilcoxon), providing no evidence of an overall gain. \Cref{sec:res-thinking} reports the counts of statistically significant changes for individual models.

Under single-shot RAG, the \DFThinkRagNQwenThree{} Qwen3 models have a mean change in Correct Abstention Rate of \DFThinkRagMeanQwenThree{}\% points when thinking is enabled, versus \DFThinkRagMeanOther{}\% points for the other \DFThinkRagNOther{} models. Without retrieval, the corresponding changes are \DFThinkNoRagMeanQwenThree{} and \DFThinkNoRagMeanOther{}\% points. These are descriptive family averages.

\paragraph{What happens to abstentions when thinking is enabled.} Each of the \DTNoRagModels{} models with switchable thinking answered every unanswerable question twice with the same weights, thinking disabled and enabled. Each answer carries the response judge's label (abstain, deny, answer). For each model and retrieval condition we cross-tabulate the two labels of each question, leaving out questions where either answer exhausted the token budget. The two discordant cells abstain$\to$deny (A$\to$D) and deny$\to$abstain (D$\to$A) are compared with an exact McNemar test: if thinking moved answers between the two labels equally often in both directions, A$\to$D would follow a binomial distribution with $n$ = A$\to$D + D$\to$A trials and probability $\frac{1}{2}$. The same test compares abstain$\to$answer with answer$\to$abstain (A$\to$X, X$\to$A), so that a loss of abstention through guessing is not counted as premise denial. \Cref{tab:thinking_pairs} and \Cref{fig:thinking_pairs} report the results, with significance assessed as in Appendix~\ref{sec:appendix-stats}.

Without retrieval, thinking turns \DTNoRagAtoD{} abstentions into premise denials and \DTNoRagDtoA{} premise denials into abstentions (\DTNoRagPairs{} matched questions). The excess of A$\to$D is significant for \DTNoRagSigAtoD{} of the \DTNoRagModels{} models, and no model shows a significant shift the other way; across models, the per-model net shift (A$\to$D $-$ D$\to$A, as a share of questions) averages $\DTNoRagNetMean$ points (Wilcoxon signed-rank, $p=\DTNoRagNetP$). Of the abstentions lost to thinking, \DTNoRagLostToDeny\% became premise denials and the rest attempted answers; the A$\to$X excess is significant for \DTNoRagSigAtoX{} models. The Qwen3.5 models rarely abstain without retrieval in either mode, so they contribute few pairs. With single-shot RAG the shift towards premise denial remains (\DTRagAtoD{} against \DTRagDtoA{}; significant for \DTRagSigAtoD{} models, the reverse for \DTRagSigDtoA), but for \DTRagSigXtoA{} models, mostly Qwen3, thinking also turns attempted answers into abstentions, which is why its net effect on Correct Abstention Rate with retrieval is mixed.

\begin{figure}[!htbp]
    \centering
    \includegraphics[width=\columnwidth]{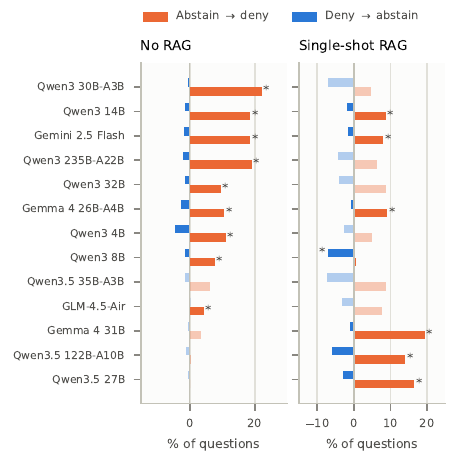}
    \caption{Matched pairs with thinking disabled and enabled on the unanswerable questions: the share of questions whose response changes from abstention to premise denial when thinking is enabled (orange) and from premise denial to abstention (blue), per model. Faded: the two counts do not differ significantly; *: they do ($q<0.05$). Models sorted by the shift without retrieval.}
    \label{fig:thinking_pairs}
\end{figure}

\begin{table*}[t]
\centering
\small
\setlength{\tabcolsep}{5pt}
\begin{tabular}{lrrrrrr}
\toprule
 & \multicolumn{3}{c}{No RAG} & \multicolumn{3}{c}{Single-shot RAG} \\
\cmidrule(lr){2-4}\cmidrule(lr){5-7}
Model & $n$ & A$\to$D / D$\to$A & A$\to$X / X$\to$A & $n$ & A$\to$D / D$\to$A & A$\to$X / X$\to$A \\
\midrule
Gemma 4 26B-A4B & 254 & \textbf{27 / 7}* & 4 / 8 & 278 & \textbf{25 / 2}* & 7 / 8 \\
Gemma 4 31B & 299 & 10 / 2 & \textbf{3 / 13}* & 300 & \textbf{58 / 3}* & 11 / 11 \\
GLM-4.5-Air & 282 & \textbf{12 / 0}* & \textbf{33 / 6}* & 299 & 23 / 10 & 28 / 26 \\
Qwen3 4B & 272 & \textbf{30 / 12}* & \textbf{34 / 11}* & 291 & 14 / 8 & \textbf{20 / 58}* \\
Qwen3 8B & 284 & \textbf{22 / 4}* & 24 / 38 & 298 & \textbf{2 / 21}* & \textbf{4 / 69}* \\
Qwen3 14B & 295 & \textbf{55 / 4}* & \textbf{37 / 15}* & 299 & \textbf{26 / 6}* & \textbf{4 / 45}* \\
Qwen3 30B-A3B & 299 & \textbf{66 / 2}* & \textbf{52 / 3}* & 299 & 14 / 21 & \textbf{4 / 44}* \\
Qwen3 32B & 294 & \textbf{28 / 4}* & 22 / 14 & 300 & 26 / 12 & 14 / 27 \\
Qwen3 235B-A22B & 300 & \textbf{57 / 6}* & 28 / 19 & 300 & 19 / 13 & \textbf{15 / 40}* \\
Qwen3.5 27B & 194 & 0 / 1 & 1 / 0 & 244 & \textbf{40 / 7}* & 11 / 5 \\
Qwen3.5 35B-A3B & 129 & 8 / 2 & 4 / 1 & 162 & 14 / 12 & 7 / 8 \\
Qwen3.5 122B-A10B & 202 & 1 / 2 & 2 / 3 & 214 & \textbf{30 / 13}* & 3 / 9 \\
Gemini 2.5 Flash & 300 & \textbf{56 / 5}* & \textbf{65 / 1}* & 300 & \textbf{24 / 5}* & \textbf{28 / 5}* \\
\bottomrule
\end{tabular}
\caption{Matched pairs with thinking disabled and enabled, per model: $n$ questions (budget-exhausted pairs left out) and the discordant counts abstain$\to$deny / deny$\to$abstain and abstain$\to$answer / answer$\to$abstain. Bold with *: the two counts differ significantly ($q<0.05$).}
\label{tab:thinking_pairs}
\end{table*}

A premise denial, including one that follows the abstention sentence requested by the prompt, does not acknowledge that the premise cannot be verified from the model's available knowledge (Appendix~\ref{sec:appendix-judges}). AbstentionBench also reports lower abstention after reasoning fine-tuning, under its separate evaluation setting~\citep{kirichenko2025abstentionbench}.

For the \DFThinkAccModels{} models with switchable thinking, mean Answer Accuracy changes by \DFThinkAccNoRag{}\% points without retrieval and \DFThinkAccRag{}\% points with single-shot RAG. \Cref{fig:thinking_acc} shows these differences per model. Thus, the change in answer quality need not have the same direction as the change in Correct Abstention Rate.

\subsection{Answer Accuracy and Correct Abstention Rate}
\label{sec:res-accuracy}
\begin{figure}[!htbp]
    \centering
    \includegraphics[width=\columnwidth]{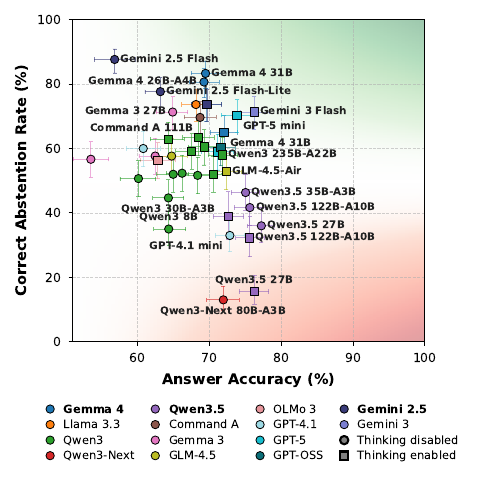}
    \caption{Correct Abstention Rate versus Answer Accuracy under single-shot RAG, one point per model and thinking setting. Colors: families; squares: thinking enabled. Bars: 95\% intervals, bootstrap over questions for accuracy and Wilson for Correct Abstention Rate. Appendix~\ref{sec:res-accuracy} reports the correlations.}
    \label{fig:all_accuracy}
\end{figure}

\begin{figure}[!htbp]
    \centering
    \includegraphics[width=\columnwidth]{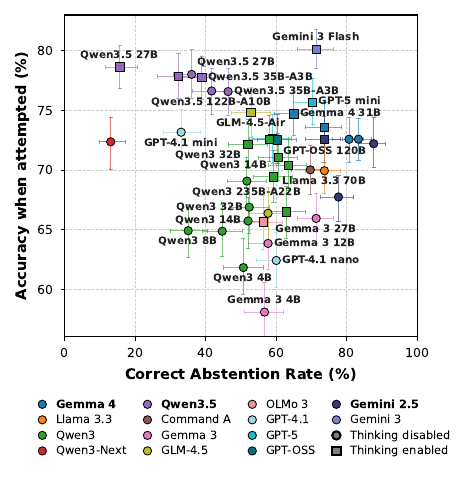}
    \caption{Accuracy when attempted (over the answerable questions a model attempted, abstentions left out) against Correct Abstention Rate, single-shot RAG. Spearman $\rho = \DCRhoSelectiveAbstain$.}
    \label{fig:selective}
\end{figure}

Under single-shot RAG, the Spearman correlation between Correct Abstention Rate and Answer Accuracy is $\rho=\DFRhoAccAbst$ ($p=\DFPAccAbst$) across the \DFNAccAbst{} model and thinking configurations (\Cref{fig:all_accuracy}). For Accuracy when attempted, the correlation is $\rho=\DFRhoAccAttAbst$ ($p=\DFPAccAttAbst$). Neither unadjusted association is significant at the 5\% level. \Cref{fig:selective} plots Accuracy when attempted against Correct Abstention Rate for each configuration.

Across the same \DFNAccAbst{} configurations, release month correlates with Answer Accuracy ($\rho=\DFRhoAccRelease$), but not with Correct Abstention Rate ($\rho=\DFRhoAbstRelease$). Partialling out release month gives a negative association between Correct Abstention Rate and Answer Accuracy ($\rho=\DFRhoAccAbstPartial$, $p=\DFPAccAbstPartial$). These cross-model comparisons do not identify the accuracy cost of changing an individual model's decision to answer: they compare different systems, and thinking configurations of the same model are not independent models. They therefore establish neither an inevitable accuracy cost nor that correct abstention is cost-free.

\subsection{Answerable Attempt Rate and Correct Abstention Rate}
\label{sec:res-tradeoff}
\begin{figure}[!htbp]
    \centering
    \includegraphics[width=\columnwidth]{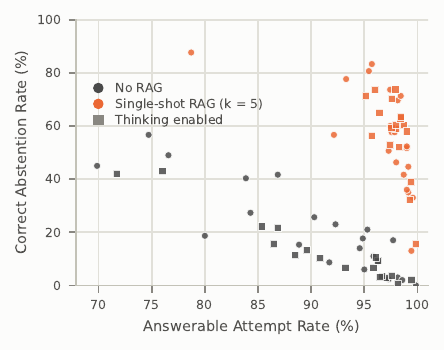}
    \caption{Correct Abstention Rate versus Answerable Attempt Rate without retrieval (gray) and with single-shot RAG (orange), one point per model and thinking setting (squares: thinking enabled). Appendix~\ref{sec:res-tradeoff} discusses the observed trade-off frontiers.}
    \label{fig:tradeoff}
\end{figure}

The trade-off frontier of a set of configurations is the highest Correct Abstention Rate observed at or above a given Answerable Attempt Rate (\Cref{fig:tradeoff} shows the configurations). It describes the evaluated systems, rather than a trade-off obtained by varying an answer threshold within one system.
In selective QA, a risk--coverage curve instead measures error among answered questions against the fraction of all questions answered as a confidence threshold varies \citep{kamath-etal-2020-selective}. Our axes are rates within separate answerability classes; neither measures error among attempted answers.

Using the \DFRagPairs{} configurations in \Cref{tab:results_rag2}, single-shot RAG yields higher Correct Abstention Rate and Answerable Attempt Rate than no retrieval in \DFTradeBothUp{} configurations. In \DFTradeCarUpAttDown{} configurations, Correct Abstention Rate is higher but Answerable Attempt Rate lower; the opposite holds for \DFTradeCarDownAttUp{}. These counts describe the directions of observed changes, not their statistical significance.

For an illustrative Answerable Attempt Rate requirement of at least \DFTradeAttMin\%, the highest Correct Abstention Rate without retrieval is \DFTradeNoRagBestCar\% (\texttt{\DFTradeNoRagBestModel}, \DFTradeNoRagBestSetting; Answerable Attempt Rate \DFTradeNoRagBestAtt\%). With single-shot RAG, it is \DFTradeRagBestCar\% (\texttt{\DFTradeRagBestModel}, \DFTradeRagBestSetting; Answerable Attempt Rate \DFTradeRagBestAtt\%). Thus, the observed trade-off frontier permits a higher Correct Abstention Rate under the same Answerable Attempt Rate requirement. The two points represent different models; they do not measure a change within one model or establish an unavoidable trade-off.

Across the \DFRagPairs{} configurations, moving from no retrieval to single-shot RAG raises mean Answerable Attempt Rate from \DFTradeNoRagAtt\% to \DFTradeRagAtt\% and mean Correct Abstention Rate from \DFTradeNoRagCar\% to \DFTradeRagCar\%. Retrieval thus improves both means in this comparison. With iterative RAG, mean Answerable Attempt Rate is \DFAttMeanIter\%, while mean Correct Abstention Rate falls to \DFMeanIter\%. \Cref{fig:scatter,fig:scatter_iterative} show individual configurations under single-shot and iterative RAG.

\subsection{Models Fine-Tuned to Abstain}
\label{sec:res-abstention-trained}
We evaluate three models fine-tuned for better abstention:

\texttt{Trust-Align 3B} and \texttt{Trust-Align 7B}~\citep{song2025trustalign}, trained with preference optimization to answer only from retrieved documents and to abstain otherwise.

\texttt{Abstain-R1 3B}~\citep{zhai2026abstainr1}, trained with reinforcement learning to answer ``I don't know'' and state what information is missing. 

Each is compared with the baseline model it was fine-tuned from, \texttt{Qwen2.5-3B-Instruct} or \texttt{Qwen2.5-7B-Instruct}. \texttt{Abstain-R1} was trained with its own prompt template, so we evaluate it with that template; the other models answer with our prompt. All models answer the same 1,000 questions as the rest of the study and are labeled and scored exactly as the other models. We report single-shot RAG.

\paragraph{How we ran the models.} We used the released Hugging Face checkpoints \texttt{declare-lab/trustalign\_qwen2.5\_3b} and \texttt{declare-lab/trustalign\_qwen2.5\_7b} (revisions \texttt{9ce97fa} and \texttt{46f3a92}) and \texttt{leoleung04/Abstain-R1} (revision \texttt{a02d6ed}), with \texttt{Qwen/Qwen2.5-3B-Instruct} and \texttt{Qwen/Qwen2.5-7B-Instruct} as the baseline models. All five are served with vLLM~\citep{kwon2023vllm} with a 32,768-token context, and decode greedily at temperature 0 with the same 16,000-token output budget, retrieval and labeling as the other models. \texttt{Abstain-R1} answers with its own prompt template, reproduced verbatim from \citet{zhai2026abstainr1}. Our setting differs from the one each method was trained for. Trust-Align learns to abstain whenever the retrieved documents do not state the answer, whereas our benchmark also accepts answers from the model's own knowledge, and \texttt{Abstain-R1} was trained without retrieved context. The comparison therefore measures how these methods carry over to our setting, not how well they perform on the tasks they were designed for.

\paragraph{Fine-tuning trades attempts for abstention along the same trade-off frontier.} Fine-tuning moves every model up and to the left in \Cref{fig:abstention_tradeoff}: it trades Answerable Attempt Rate for Correct Abstention Rate. Fine-tuning raises CAR from \DAQThreeRagCar\% to \DATAThreeRagCar\% for \texttt{Trust-Align 3B} and from \DAQSevenRagCar\% to \DATASevenRagCar\% for \texttt{Trust-Align 7B}, but these models attempt only \DATAThreeRagAtt\% and \DATASevenRagAtt\% of answerable questions respectively, against at least \DAMainAttMin\% for every other configuration. With its own template, \texttt{Abstain-R1 3B} reaches \DAARoneRagNativeCar\% at an attempt rate of \DAARoneRagNativeAtt\%, which lies on the trade-off frontier of the other configurations: \texttt{\DAFrontierModel} matches this Correct Abstention Rate at a higher attempt rate (\DAFrontierAtt\%). Part of this shift comes from the template itself: run with the same template, the baseline model (\texttt{Qwen2.5-3B-Instruct}) reaches \DAQThreeRagNativeCar\% at \DAQThreeRagNativeAtt\%. Trust-Align exceeds the trade-off frontier's highest Correct Abstention Rate (\DAFrontierCar\%) only at attempt rates no other configuration goes down to, and by less than the uncertainty of its own estimate (95\% interval \DATAThreeRagCarLo--\DATAThreeRagCarHi\% for the 3B model). Fine-tuning for abstention therefore moves models along the trade-off rather than beyond it.

\paragraph{Models fine-tuned for abstention have lower Answer Accuracy.} Answer Accuracy falls from \DAQThreeRagAcc\% to \DATAThreeRagAcc\% for \texttt{Trust-Align 3B} and from \DAQSevenRagAcc\% to \DATASevenRagAcc\% for \texttt{Trust-Align 7B} (\Cref{fig:abstention_accuracy}). \texttt{Abstain-R1 3B} scores \DAARoneRagNativeAcc\%, below its baseline model (\DAQThreeRagAcc\%). Much of the fall comes from abstaining on answerable questions that other models answer correctly. On the \DATAThreeDeclinedN{} answerable questions on which \texttt{Trust-Align 3B} abstains, the other configurations average \DATAThreeDeclinedMainAcc\% Answer Accuracy (\DAMainAccAll\% over all answerable questions), and its baseline model \DATAThreeDeclinedBaseAcc\%; for \texttt{Trust-Align 7B} the figures are \DATASevenDeclinedN{} questions, \DATASevenDeclinedMainAcc\% and \DATASevenDeclinedBaseAcc\%. On the questions they do attempt, both Trust-Align models are about as accurate as their baseline models (\DATAThreeRagAccAtt\% against \DAQThreeRagAccAtt\%, and \DATASevenRagAccAtt\% against \DAQSevenRagAccAtt\%).

\subsection{Qualitative Examples: Unsupported Answers and Incorrect Abstentions}
\label{sec:qualitative}

We illustrate two failure modes: an unsupported answer to an unanswerable question and an abstention on an answerable question. We say \emph{confabulation} rather than \emph{hallucination} for an answer that fills a knowledge gap with plausible but invented details the model does not recognize as wrong, following \citet{smith2023confabulation}. The first comes from \texttt{GPT-OSS 120B} and the second from \texttt{Gemini 2.5 Flash} with thinking disabled; bold marks the error.

\noindent
\textbf{Confabulation Example:}
\begin{itemize}[leftmargin=*]
\item \textit{Question: I am flying with my Elytra and need to boost my speed, so I aim a firework rocket at my feet. What are the distinct archetypes of the Sulfur Cube entity?}
\item \textit{LLM Answer (no RAG): ``The Sulfur Cube has \textbf{three separate `archetypes'} (variants) that differ in behavior and strength'', listed as \textbf{Normal}, \textbf{Charged} (``created when a Normal Sulfur Cube is struck by lightning'') and \textbf{Explosive}.}
\item \textit{LLM Answer (single-shot RAG): ``The provided context does not contain the answer. I would need to know more about the Sulfur Cube entity and its archetypes to respond.''}
\item \textit{LLM Answer (iterative RAG): ``The \textbf{Magma Cube (sometimes called a `Sulfur Cube')} comes in \textbf{three size-based archetypes}'': large, medium and small.}
\item \textit{Analysis:} The Sulfur Cube was added after the corpus cut-off. It has twelve archetypes, and all are set by the block it absorbs. From memory alone, the model invents three. With one round of retrieval, the $C_\text{pre}$ pages it receives say nothing about the mob, and it abstains correctly. After further rounds it retrieves the \textit{Magma Cube} page and confabulates that the two mobs are the same. This example illustrates a failure consistent with the aggregate comparison in \Cref{sec:retrieval-results}.
\end{itemize}

\noindent
\textbf{Incorrect Abstention Example:}
\begin{itemize}[leftmargin=*]
\item \textit{Question: I am sneaking on a slime block to avoid bouncing while a cave spider attacks me with poison. Which mob would I need to find to make the attacking cave spider run away?}
\item \textit{LLM Answer (single-shot RAG): ``The provided context does not contain the answer. \textbf{I would need to know what mobs scare cave spiders.}''}
\item \textit{Analysis:} The top retrieved passage, from the \textit{Cave Spider} page, lists among the mob's behaviors that cave spiders ``are scared of armadillos''. The model abstains although the answer is in its context. \texttt{Gemini 2.5 Flash} without thinking attempts only \DCAttGeminiTwoFiveFlashAPIoff\% of answerable questions under single-shot RAG.
\end{itemize}

\section{Answering Only from the Context}
\label{sec:appendix-external}
Context-grounded benchmarks such as RefusalBench~\citep{muhamed2025refusalbench} instruct the model to answer only from the supplied context. We tested RefusalBench's answer prompt on a sample of \DENatUnans{} unanswerable and \DENatAns{} answerable \name{} questions, each given its single top retrieved chunk, with \DEConfigs{} configurations of \DEModels{} open-weight models; a separate judge, an LLM that sees only the response, decides whether it declines. Under this prompt, the configurations decline \DENatRBMin--\DENatRBMax\% of the unanswerable questions, close to the ceiling, but they also decline \DEAnsRefRBMin--\DEAnsRefRBMax\% of the answerable ones. Under our prompt, which lets models use their own knowledge, they decline \DEAnsRefDCMin--\DEAnsRefDCMax\% of the answerable questions. The retrieved chunk is the one an answerable question was written from for only \DEGoldHit\% of these questions, so a model that follows a context-only instruction must decline many answerable questions whatever it knows. Under such an instruction, the task reduces to deciding whether the answer is in the retrieved chunk. This is why \name{} lets models use their own knowledge and counts an abstention as correct only when neither the context nor the model's knowledge can answer the question.

\section{Examples of Unanswerable Questions}
\label{sec:appendix-unasw}

Designing questions whose answerability is certain can be challenging, even for humans. Below is an example illustrating how temporal shifts create verifiably unanswerable scenarios.

\paragraph{Unanswerable Scenario}
Consider a RAG system posed with the following question: \textit{``I ride my horse across the plains, holding the jump control to charge a high leap over the fence gate blocking my path. If I equip copper horse armor, how many armor points will it provide?''}

Copper horse armor was added to \textit{Minecraft} after the corpus cut-off: its wiki page was created on 29 July 2025, when Java Edition snapshot 25w31a was released~\citep{minecraft2025snapshot25w31a}. The page exists only in $C_\text{post}$, so the retriever cannot reach it, and the correct response is to abstain. This makes abstention difficult: the scenario leads the retriever to closely related $C_\text{pre}$ pages about horses, horse armor and leather, iron, gold and diamond horse armor, each with its own armor value. The model must recognize that none of them covers copper, rather than substituting a neighboring value. \Cref{tab:example} shows real answers of each kind.

\paragraph{Answerable Scenario}
Consider a RAG system posed with the following question: \textit{``I am breeding pandas in the bamboo jungle, hoping to get a playful one instead of a lazy or angry panda. How do the main and hidden genes determine whether my panda becomes playful or a recessive type like weak or brown?''}

Reference answer (abridged): \textit{``A panda's personality is set by its main gene when that gene is dominant, so it is playful whenever its main gene is playful, whatever its hidden gene. The recessive weak and brown types appear only when both genes carry that trait.''}

The answer is on the \textit{Panda} page in $C_\text{pre}$. The model must apply the dominance rule to both genes rather than assume that either gene alone decides the trait.

\providecommand{\checkmark}{\ensuremath{\surd}}

\section{Corpus Construction}
\label{sec:appendix-scraping}

Both corpora come from the \textit{Minecraft} Wiki (\url{https://minecraft.wiki}) through the MediaWiki API (\texttt{api.php}). The corpus cut-off is 31 March 2025, 23:59:59 UTC.

\paragraph{$C_\text{pre}$ (retrievable).} A breadth-first crawl starts from 36 seed pages (such as Trading, Brewing, Enchanting, Crafting, Smelting, Mob, Item, Block, ...) and follows linked pages. For every page it stores the \emph{latest revision at or before the corpus cut-off}. 

We exclude version and changelog pages, pages on live events, mob votes, skin packs, MINECON, films, mob-vote proposal pages, the Mentioned features page (features announced but never added), and pages shorter than 200 characters after cleaning.

The corpus has 2,642 pages, with revision dates from 22 October 2022 to 31 March 2025.

\paragraph{$C_\text{post}$ (questions only).} A crawl starts from 35 titles naming post-cut-off content (the 2025--2026 game drops \textit{Chase the Skies}, \textit{The Copper Age}, \textit{Mounts of Mayhem}, \textit{Tiny Takeover} and \textit{Chaos Cubed}, and items such as copper tools and armor, sulfur and cinnabar blocks) and follows links. A page is kept only if its \emph{first} revision is after the corpus cut-off, so a page that existed before the corpus cut-off and was later updated is excluded. 

Edition and version pages, spin-off games (by title), media, tutorials and pages already in $C_\text{pre}$ are excluded, as are pages under 200 characters. $C_\text{post}$ has 170 pages, created between 2 April 2025 and 6 September 2026, stored at their latest revision when scraped (8 September 2026).

\paragraph{Cleaning and indexing.} Wikitext is reduced to text: templates are replaced by their last argument, links by their display text, and galleries, comments, file embeds and interlanguage links are removed; tables are kept as wikitext. Sections about other versions or out-of-game material are dropped: History, Trivia, Data values, Sounds, Achievements, Advancements, Screenshots, Navigation, See also, References, External links, In other media, Mob votes, Upcoming, Joke, Unimplemented and Removed mobs and features, April Fools, and Mentioned. 

Pages are split recursively into chunks of at most 1,500 characters with 250 characters of overlap, each prefixed with \texttt{[Page: <title>]}: 11,612 chunks for $C_\text{pre}$ and 712 for $C_\text{post}$. Chunks are embedded with \texttt{Qwen3-Embedding-8B (Q8\_0)}~\citep{zhang2025qwen3embedding} and stored in ChromaDB; a BM25 index~\citep{robertson2009bm25} is built over $C_\text{pre}$ only, with our implementation's defaults $k_1{=}1.5$ and $b{=}0.75$. The retrievers built on this index are described in \Cref{sec:appendix-retrieval}.

\paragraph{Audit.}
After generation, we reviewed every unanswerable question by hand and removed 21 questions with malformed facts, such as a spin-off game (``Coral Rise''), fabricated items (``steel and fire striker'') and impossible scenarios (``creeper wielding a diamond sword''). Questions whose answer can only be guessed by analogy with an older, general mechanic (e.g.\ whether a new copper tool accepts an enchantment its iron counterpart accepts) were kept: assuming that an old rule carries over to a new item is the behavior the benchmark measures, and such a guess may be wrong. Each removed question was replaced by an unused candidate from the same build that had passed the same checks. The benchmark therefore keeps 700 answerable and 300 unanswerable questions, and every model answered the replacements under every condition.

\section{Retrieval Pipelines}
\label{sec:appendix-retrieval}

Every pipeline searches the $C_\text{pre}$ index with the full query (scenario and question) and gives the model $k$ chunks ($k{=}5$ unless stated). The default is hybrid retrieval; the retrieval ablation compares all six.

\begin{itemize}[leftmargin=*]
\item \textbf{Dense.} The query is embedded with \texttt{Qwen3-Embedding-8B}, and the $k$ chunks with the highest cosine similarity are returned~\citep{karpukhin-etal-2020-dense,zhang2025qwen3embedding}.
\item \textbf{BM25.} Okapi BM25 over lower-cased word tokens with English stopwords removed, $k_1{=}1.5$, $b{=}0.75$~\citep{robertson2009bm25}. It matches exact terms, such as item names, that an embedding may blur.
\item \textbf{Hybrid (default).} Dense retrieval and BM25 each return their top 20 chunks, and the two rankings are merged by reciprocal rank fusion (RRF)~\citep{cormack2009rrf}: a chunk scores $\sum_l 1/(60 + r_l)$, where $r_l$ is its rank in list $l$, and the $k$ highest-scoring chunks are kept. RRF uses ranks rather than raw scores, so cosine similarities and BM25 scores need no common scale, and a chunk ranked highly by both retrievers comes first.
\item \textbf{Hybrid + rerank.} The 20 highest fused chunks are rescored by a cross-encoder, \texttt{llama-nemotron-rerank-1b-v2}, which reads the query and each chunk together~\citep{nogueira2019passage,nvidia2025nemotronrerank}; the $k$ highest are kept.
\item \textbf{Hybrid + rewrite.} An LLM (\texttt{Qwen3 30B-A3B}, thinking disabled) rewrites the query into three short search queries that target the question rather than the scenario~\citep{ma-etal-2023-query}. The original query and the three rewrites each give a dense and a BM25 list, and all eight lists are fused by RRF. Rewrites are generated once and cached, so every model receives the same ones.
\item \textbf{Hybrid + rerank + rewrite.} Rewriting and fusion as above, then cross-encoder reranking of the 20 highest fused chunks against the original query.
\end{itemize}
Follow-up queries in iterative RAG use hybrid retrieval (\Cref{alg:iterative}).

\section{Question Generation}
\label{sec:appendix-generation}

Both halves are written by the same three stages (Algorithms~\ref{alg:draft}, \ref{alg:verify} and~\ref{alg:assemble}); the only difference is the corpus (pre- or post-cut-off) the target chunk comes from. Appendix~\ref{sec:appendix-prompts-generation} gives the prompts, and Appendix~\ref{sec:appendix-unasw} shows an unanswerable and an answerable example. Dense retrieval (\texttt{Qwen3-Embedding-8B}) returns the ten chunks of that corpus most similar to the scenario, excluding technical pages (engine internals, rendering and data formats) and, for answerable questions, the scenario's own source chunks. The target chunk is drawn at random from the five highest-ranked chunks that remain. The fact chosen from this target chunk must not be stated by: the scenario's three source chunks for answerable questions, so that the scenario does not give the answer away; and the three $C_\text{pre}$ chunks closest to the target for unanswerable questions, so that the fact is new relative to the pre-cut-off corpus. The generator is \texttt{Qwen3.5 122B-A10B} (8-bit, thinking disabled; temperature 0.3 for the scenario, 0.5 for the question, 0 otherwise). The verifier is one judge, \texttt{Gemma 4 31B} (8-bit, thinking disabled), which reads $C_\text{pre}$ through hybrid retrieval with reranking (10 chunks per query). A verdict that cannot be parsed rejects the candidate. Each stage runs over the whole benchmark at once, so each model is loaded once.

\paragraph{Cue filter.} A question is rejected if it names its own novelty or repeats release talk from the source passage: \textit{wonder(ed/ing), suddenly, notice(d), heard, announced, announcement, new(ly), recent(ly), latest, update(d), added, introduced, change(d), no longer, nowadays, these days, now that}; \textit{Minecraft Live, snapshot, pre-release, beta build}; any year 20xx; any month name. The words are not shown to the generator: listing them in the prompt made the wording stiffer and easier to classify (style-only AUC 0.73 vs.\ 0.64 in a pilot).

\paragraph{Wording separability (style AUC).}\label{sec:appendix-style} If the two halves were worded differently, a model could flag unanswerable questions by their phrasing rather than by what it knows. We test this with a bag-of-words classifier that predicts a question's half from its text alone: TF-IDF features over words of at least three letters that occur in at least three questions, an L2-regularized logistic regression, and ROC AUC under 10-fold cross-validation (0.5 is chance, 1 is perfect separation). The halves legitimately differ in what they ask about, since unanswerable questions name post-cut-off items, so the \emph{style-only} AUC first removes every word that occurs in the title of a $C_\text{pre}$ or $C_\text{post}$ page (plurals folded), leaving the phrasing; novelty words such as ``new'' or ``update'' are never removed, so they would still show. On the final benchmark, the style-only AUC is \DMStyleFinal{} (\DMStyleFinalFull{} on the full text, where item names separate the halves).

\begin{algorithm}[t]
\caption{Draft one candidate question for half $h$.}
\label{alg:draft}
\small
\begin{algorithmic}[1]
\Require $h \in \{\textsc{answerable}, \textsc{unanswerable}\}$
\State $S_{\text{ctx}} \gets$ 3 chunks drawn uniformly from $C_{\text{pre}}$
\State $S \gets G_{\text{scenario}}(S_{\text{ctx}})$ \Comment{$\approx$30-word scenario}
\State $X \gets C_{\text{pre}}$ if $h = \textsc{answerable}$ else $C_{\text{post}}$
\State $T \gets R_X(S, 10)$ minus technical pages (and minus $S_{\text{ctx}}$ if answerable)
\State \textbf{if} $T = \emptyset$ \textbf{then reject}
\State $t \gets$ uniform choice among the top 5 of $T$
\State $O \gets S_{\text{ctx}}$ if answerable else $R_{C_{\text{pre}}}(t, 3)$ \Comment{what the player has}
\State $F \gets G_{\text{facts}}(O, t)$ \Comment{facts $t$ states that $O$ does not}
\State \textbf{if} $F = \emptyset$ \textbf{then reject}
\State $f \gets$ uniform choice from $F$;\ $q \gets G_{\text{question}}(S, f, t)$
\State \textbf{if} $S \oplus q$ is not one question with one `?' or exceeds 55 words, or $q$ matches the cue filter \textbf{then reject}
\State \Return $(S, t, f, S \oplus q)$
\end{algorithmic}
\end{algorithm}

\begin{algorithm}[t]
\caption{Verify a candidate. The checks are mirrored, so each half is held to the condition that makes its label true.}
\label{alg:verify}
\small
\begin{algorithmic}[1]
\Require candidate $(S, t, f, Q)$, half $h$
\State $D_Q \gets R^{+}_{C_{\text{pre}}}(\textsc{asked}(Q)) \cup R^{+}_{C_{\text{pre}}}(Q)$ \Comment{10 chunks per query}
\If{$h = \textsc{answerable}$}
  \State \textbf{if} $\textsc{Stated}(f, S)$ \textbf{then reject} \Comment{scenario gives it away}
  \State \textbf{if not} $\textsc{Answers}(Q, f, D_Q)$ \textbf{then reject}
\Else
  \State \textbf{if} $\textsc{Stated}(f, R^{+}_{C_{\text{pre}}}(f))$ \textbf{then reject} \Comment{old fact}
  \State \textbf{if} $\textsc{Answers}(Q, f, D_Q)$ \textbf{then reject} \Comment{drifted}
\EndIf
\State \Return \textbf{accept}
\end{algorithmic}
\end{algorithm}

\begin{algorithm}[t]
\caption{Assemble the final set: chunk cap $\kappa{=}2$, page cap $\pi{=}10$.}
\label{alg:assemble}
\small
\begin{algorithmic}[1]
\Require verified candidates $V$, target $n$
\State $A \gets \emptyset$
\ForAll{$x \in V$ while $|A| < n$}
  \State \textbf{if} chunk $t_x$ used $\kappa$ times \textbf{or} its page $\pi$ times \textbf{then skip}
  \State $A \gets A \cup \{x\}$
\EndFor
\State \Return $A$
\end{algorithmic}
\end{algorithm}

\noindent $R_X(q,k)$ is dense retrieval over corpus $X$; $R^{+}$ is hybrid retrieval with reranking and is used only by the verifier. $G_\ast$ are generator calls and \textsc{Stated}, \textsc{Answers} are verifier calls (prompts in \Cref{sec:appendix-prompts}). \textsc{asked}$(Q)$ is the question sentence without the scenario. Nothing selects for whether the answer chunk is easy to retrieve; that rate is measured instead (the answer chunk is among the top 5 retrieved for \DCRecallHitFive\% of answerable questions).

\DCappendixsection{Choice of Models for Building and Scoring}
\label{sec:appendix-choices}
Each step that builds or scores the benchmark uses a different model from the step it checks, so that no single model verifies its own questions or alone scores accuracy for itself (\Cref{tab:components}). The specific models were chosen by the comparisons below, run while the pipeline was designed.

\paragraph{Generator.} We ran the same scenario-first pipeline, seed and verifier (\texttt{Laguna S 2.1}) with three generators, each targeting \DMTarget{} answerable and \DMTarget{} unanswerable questions (\Cref{tab:generators}). The verifier rejected unanswerable candidate questions whose fact $C_\text{pre}$ already states at similar rates for all three (\DMKnownQwenThreeTwoThreeFive--\DMKnownGemmaFour\%). \texttt{Qwen3.5 122B-A10B} was chosen because a classifier separates its answerable from its unanswerable questions least well by wording alone (style-only AUC \DMStyleQwenThreeFive{}, where 0.5 is chance; classifier in Appendix~\ref{sec:appendix-style}), so the two halves cannot be told apart by phrasing; \texttt{Qwen3 235B-A22B} lost \DMLostQwenThreeTwoThreeFive{} drafts to unparsable output and its questions were the easiest to separate (AUC \DMStyleQwenThreeTwoThreeFive).

\paragraph{Verifier.} With \texttt{Qwen3.5 122B-A10B} generating, \texttt{Laguna S 2.1} rejected none of \DMVerLagunaDrafts{} answerable drafts as unsupported by $C_\text{pre}$, whereas \texttt{Gemma 4 31B} rejected \DMVerGemmaRej{} of \DMVerGemmaDrafts{} (\DMVerGemmaRejPct\%) in a pilot run; the retriever found the source chunk of \DMVerGemmaHitFive\% of the answerable questions Gemma accepted, against \DMVerLagunaHitFive\% for Laguna's. We therefore use \texttt{Gemma 4 31B}, the stricter verifier. In an earlier build over the same $C_\text{post}$ (with $C_\text{pre}$ still holding the proposal pages later removed), three verifiers from different families flagged similar shares of \DMPanelFacts{} candidate facts as already stated in $C_\text{pre}$: \texttt{Qwen3.5 122B-A10B} \DMPanelQwenThreeFive\%, \texttt{Gemma 4 26B-A4B} \DMPanelGemmaFour\% and \texttt{GLM-4.5-Air} \DMPanelGLM\%.

\paragraph{Judges.} The response judge, \texttt{Gemma 4 26B-A4B}, is checked against two judges from other families (\Cref{sec:appendix-judges}). Answer Accuracy is the mean of three accuracy judges from different families, so no answer is scored only by a model of its own family. \Cref{tab:judges} reports their agreement, and \Cref{sec:res-judge-selfpref} tests for self-preference. Both are compared with human annotators in \Cref{sec:appendix-human}.

\begin{table}[!htbp]
\centering
\small
\resizebox{\columnwidth}{!}{%
\begin{tabular}{@{}lrrrrrr@{}}
\toprule
Generator & Accepted & Lost to & Fact in & Style & Hit@5 & CAR \\
 & (ans. / unans.) & format & $C_\text{pre}$ (\%) & AUC & (\%) & (\%) \\
\midrule
\texttt{Qwen3.5 122B-A10B} & 100 / 82 & 35 & 18.0 & 0.52 & 80.0 & 84.1 \\
\texttt{Qwen3 235B-A22B} & 100 / 78 & 194 & 17.3 & 0.65 & 91.0 & 82.1 \\
\texttt{Gemma 4 31B} & 100 / 87 & 33 & 20.2 & 0.57 & 87.0 & 85.1 \\
\bottomrule
\end{tabular}
}
\caption{Generator comparison: the same pipeline, seed and verifier (\texttt{Laguna S 2.1}) with three generators, \DMTarget{} answerable and \DMTarget{} unanswerable targets each. \emph{Accepted}: questions that passed every check. \emph{Lost to format}: drafts with no parsable fact list or question. \emph{Fact in $C_\text{pre}$}: unanswerable drafts rejected because $C_\text{pre}$ already states the fact. \emph{Style AUC}: how well a classifier on wording features separates answerable from unanswerable questions (0.5 is chance). \emph{Hit@5}: answerable questions whose source chunk the default retriever returns in its top five. \emph{CAR}: Correct Abstention Rate of \texttt{Gemma 4 31B} (8-bit, thinking disabled) with single-shot RAG, labeled by the same response judge as the main results.}
\label{tab:generators}
\end{table}

\DCappendixsection{Evaluation Setup}
\label{sec:appendix-setup}\label{sec:appendix-iterative-rag}

Every model answers all 1,000 questions under each condition at temperature 0, with thinking enabled and disabled where switchable. Open-weight models were run at the highest weight precision possible on our servers (\Cref{tab:models}). Appendix~\ref{sec:appendix-models} lists the models, Appendix~\ref{sec:appendix-ans-gen} gives the answer prompts, and Appendix~\ref{sec:appendix-results} reports the results for every model. The output budget is 16,000 tokens including thinking; an answer that uses it up without a final answer is \emph{budget-exhausted}: it is excluded from attempt and abstention rates and scores 0 in Answer Accuracy. Local models are served with vLLM~\citep{kwon2023vllm}; \Cref{tab:models} lists them. Closed models and \texttt{GPT-OSS 120B} are queried through OpenRouter under each retrieval condition. OpenRouter does not forward a temperature to \texttt{GPT-5 mini} and \texttt{GPT-5 nano}, so they run at their default settings. \texttt{GPT-OSS 120B} always thinks, at the default reasoning effort~\citep{openai2026modelpages,openai2025gptoss}. \texttt{Gemini 2.5 Flash}, which thinks by default~\citep{google2026geminithinking}, is run with thinking explicitly enabled and explicitly disabled. \texttt{Gemini 3 Flash (preview)}~\citep{google2025gemini3flash} is run at its low thinking level, to bound cost. \texttt{GPT-4.1 mini}, \texttt{GPT-4.1 nano} and \texttt{Gemini 2.5 Flash-Lite} are called without a thinking setting.

\paragraph{Statistical tests.}\label{sec:appendix-stats} All tests are two-sided at the 5\% level. \emph{Paired comparisons of two runs on the same questions} (single-shot RAG versus no retrieval, iterative versus single-shot RAG, thinking enabled versus disabled) use the exact McNemar test~\citep{mcnemar1947} on whether each unanswerable question is answered with \textsc{abstain}: a binomial test on the questions where the two runs disagree. Questions on which either run exhausted its output budget are left out of the pair. Each family of comparisons is corrected with the Benjamini--Hochberg procedure~\citep{benjamini1995}, and a difference is called significant at $q<0.05$: all paired comparisons of the main grid at temperature 0 form one family, the retrieval-pipeline ablation forms another, and the answer changes when thinking is enabled (\Cref{sec:res-thinking-details}) form one family per retrieval condition and kind of change. \emph{Associations across models} (Correct Abstention Rate against total or active parameters, release date and Answer Accuracy) use Spearman's $\rho$ over one point per model, per family or per configuration, as stated with each result, with the $p$-value from the $t$ approximation; these are single tests and are not corrected for multiplicity. The association between Answer Accuracy and Correct Abstention Rate with release month held fixed is the Pearson correlation of the rank residuals after regressing each variable's ranks on the release month's ranks. \emph{Mean changes across models}, such as the effect of enabling thinking, use the Wilcoxon signed-rank test~\citep{wilcoxon1945} on the per-model differences. \emph{Intervals} are 95\%: Wilson intervals for rates; for Answer Accuracy, a bootstrap over questions (1,000 resamples of the per-question mean, since every run answers the same questions); and for means over models, a $t$ interval.

\paragraph{Iterative RAG.} Algorithm~\ref{alg:iterative} extends single-shot retrieval with self-reflection, after Self-RAG~\citep{asai2023selfrag} and the query-level self-evaluation of \citet{303de1805f3931282381c71f804bae498f57ade0}. The model under test judges whether its context suffices (reflection prompt in \Cref{sec:appendix-prompts}); if not, its follow-up queries share the original budget of $k$ chunks.

\begin{algorithm}[t]
\caption{Iterative RAG ($k{=}5$, at most $m{=}2$ extra rounds).}
\label{alg:iterative}
\small
\begin{algorithmic}[1]
\Require question $q$, model $M$
\State $C \gets R_{C_\text{pre}}(q, k)$
\For{$i = 1, \dots, m$}
  \State $(a, Q') \gets M_\text{reflect}(q, C)$ \Comment{JSON; $|Q'| \le 5$}
  \State \textbf{if} $a \ne$ \textit{Insufficient} \textbf{or} $Q' = \emptyset$ \textbf{then break}
  \State $C \gets C \cup \bigcup_{q' \in Q'} R_{C_\text{pre}}(q', \max(1, \lfloor k/|Q'| \rfloor))$ \Comment{dedup.\ by chunk}
\EndFor
\State \Return $M_\text{answer}(q, C)$
\end{algorithmic}
\end{algorithm}

\section{Labeling and Scoring}
\label{sec:appendix-judges}

\paragraph{Response labels.} Every response is labeled by one response judge, \texttt{Gemma 4 26B-A4B} (thinking disabled), in a single pass with one prompt (\Cref{sec:appendix-prompts-judges}). The judge sees the question and the response, with any thinking trace removed, and assigns exactly one of three labels: \textsc{abstain} (the response does not answer, for example because it says that it does not know or that the context does not contain the answer, and neither denies the item nor answers anyway), \textsc{deny} (it states that the asked-about item does not exist) or \textsc{answer} (it answers the question, treating the item as real). Responses longer than \DCRespJudgeCapChars{} characters are truncated to that length (\DCRespJudgeLongPct\% of responses). Responses that exhaust the output budget without a final answer are not labelled (\Cref{sec:appendix-setup}). Two judges from other families, \texttt{Qwen3.8 27B} and \texttt{Llama 3.3 70B}, relabel the same \DCCrossQwenN{} answers with the same prompt, sampled equally across Gemma's three labels and the three retrieval conditions. They agree with Gemma on \DCCrossQwenAgree\% and \DCCrossLlamaAgree\% of the answers (Cohen's $\kappa$ = \DCCrossQwenKappa{} and \DCCrossLlamaKappa). Most disagreements are responses that give the abstention sentence requested by the prompt and then a premise denial, which the other judges often read as abstentions. The next paragraphs explain why such premise denials do not count as correct abstentions.

\paragraph{Acknowledged uncertainty and premise denial.} Correct Abstention Rate counts only responses labeled \textsc{abstain}: the model acknowledges missing knowledge without a premise denial or an unsupported answer. A \textsc{deny} response asserts, without support, that the asked-about item or feature does not exist; this includes a premise denial followed by a substitution or a guessed answer. The available sources do not establish whether the item exists when the question is asked, so a premise denial does not acknowledge this knowledge limit, even if it was true at the corpus cut-off.

\citet{kim-etal-2021-linguist} word such abstentions as ``we could not verify that\ldots'' rather than asserting that an unverifiable presupposition is false, because the fact may hold outside the available source. \citet{kim-etal-2023-qa} note that an assumption's status can change over time and that non-existence is often inferred from omission, and only a heuristic. A premise denial about a post-cut-off item makes the same inference from the retrieved chunks or the corpus the model was trained on.

AbstentionBench likewise treats acknowledging the knowledge cut-off as the appropriate response to questions whose answers changed after it~\citep{kirichenko2025abstentionbench}. Reframing the question is also distinct from abstaining~\citep{wen-etal-2025-know}. On answerable questions, only \textsc{abstain} counts as not attempting; a premise denial there counts as an attempted, incorrect answer, since every item asked about exists.

\paragraph{Response categories and alternative counting rules.} \Cref{fig:outcomes} splits each family's responses to unanswerable questions into the three labels. Correct Abstention Rate is the share labeled \textsc{abstain} (the strict rule). A lenient rule, as in false-premise benchmarks, would also count \textsc{deny} as a correct abstention. Such a rule measures how often models decline, not the knowledge-gap recognition our metric requires. Over all configurations, the lenient rule gives mean rates of \DCCautionNoRagLenient\% without retrieval, \DCCautionRagLenient\% with single-shot RAG and \DCCautionIterLenient\% with iterative RAG, against \DCCautionNoRagStrict\%, \DCCautionRagStrict\% and \DCCautionIterStrict\% under the strict rule. Some families give premise denials much more often than others, which is why we use the strict rule. With single-shot RAG, Command~A abstains on \DCFamCommandARagStrict\% of unanswerable questions and Qwen3.5 on only \DCFamQwenThreeFiveRagStrict\%, because Qwen3.5 responds with premise denial on most of the others. Under the lenient rule, the two families would have similar rates (\DCFamCommandARagLenient\% and \DCFamQwenThreeFiveRagLenient\%), although Command~A acknowledges that it does not know far more often. The rule also changes two findings. Under the lenient rule, single-shot RAG still raises the rate significantly for \DFRagUpLenient{} of \DFRagPairsLenient{} configurations and iterative RAG still lowers it for \DFItDownLenient{}. However, enabling thinking without retrieval no longer lowers it (mean change \DFThinkNoRagMeanLenient\% points; \DFThinkNoRagDownLenient{} significant decreases and \DFThinkNoRagUpLenient{} significant increases), and release month becomes positively associated with it ($\rho=\DFRhoReleaseModelLenient$, $p=\DFPReleaseModelLenient$ over \DFNReleaseModelLenient{} models), consistent with newer families giving more premise denials.

\paragraph{Answer accuracy.} Reference answers come from an oracle model, \texttt{DeepSeek V4.1 Flash}, given the full wiki pages each question was written from and the scenario chunks, without web search (the oracle model's prompt still contains a sentence about web search, but no search tool was provided). Three accuracy judges from different families (\texttt{Gemma 4 31B}, \texttt{Qwen3.8 27B} and \texttt{Laguna S 2.1}) score every answer 1, 0.5 or 0 against the reference answer. An abstention scores 0, and a run's accuracy is the mean over the three accuracy judges. The leakage audit scores no-RAG answers to unanswerable questions in the same way against reference answers about the current game.

\paragraph{How we account for class imbalance.}\label{sec:appendix-class-balance} Correct Abstention Rate (CAR) is computed only on the 300 unanswerable questions and Answerable Attempt Rate (AAR) only on the 700 answerable ones, so neither depends on the 700:300 split. Balanced F1 combines the two rates as if both classes had equal size. It treats an attempted answer as the positive prediction, which is correct on an answerable question and wrong on an unanswerable one. Its recall is therefore AAR, the share of answerable questions attempted, and its precision $P$ is the share of attempts that fall on answerable questions when the two classes are equal in size. Since $1-\mathrm{CAR}$ is the share of unanswerable questions attempted,
\begin{gather*}
P = \frac{\mathrm{AAR}}{\mathrm{AAR} + (1-\mathrm{CAR})},\\
\text{Balanced F1} = \frac{2\,P \cdot \mathrm{AAR}}{P + \mathrm{AAR}}.
\end{gather*}
To test the stability of this balanced-F1 ranking, we draw \DBDraws{} class-balanced subsamples (300 answerable questions without replacement and 300 unanswerable questions with replacement) and rank the \DBConfigs{} configurations in each. The ranking is stable: its Spearman correlation with the full ranking lies between \DBRhoLo{} and \DBRhoHi{} for 90\% of subsamples, and \texttt{\DBTopConfig} ranks first in \DBTopShare\% of them. Configurations in the middle of the ranking are less clearly separated: across 90\% of subsamples, the median configuration's rank spans \DBRankWidthMed{} places.

\paragraph{Human validation.}
\label{sec:appendix-human}
\textit{Annotators.} The annotators are three final-year undergraduate computer science and engineering students at a university in the United States, recruited through an open offer to the first author's classmates. Each worked independently and was paid 25~USD for the task, which we estimate to take 1.5 to 2 hours (12.50 to 16.70~USD per hour, above the U.S. federal minimum wage of 7.25~USD). The items contain only \textit{Minecraft} questions and model answers, so the task carries no foreseeable risk, and no personal information was collected beyond the labels. Before starting, the annotators were told that their labels would be used in a research publication, and agreed in writing. Our institution's review board reviewed the study and determined that it does not qualify as human subjects research. The full annotation guidelines and the labeling interface are included as a zip file in the supplementary material.

The same 100 items (in two parts) were labeled by \HAAnnotators{} annotators on a local web page that showed only the question, the answer and, for accuracy, the reference answer; they saw neither the model nor any judge's label. Part~1 has \HAOneN{} answers to unanswerable questions, sampled in equal numbers from the response judge's three labels across models and retrieval conditions, each to be labeled as an abstention, a premise denial or an attempted answer. Part~2 has \HATwoN{} attempted answers to answerable questions, sampled in equal numbers from low ($\leq 0.25$), middle and high ($> 0.75$) mean scores of the accuracy judges, each to be scored 1, 0.5 or 0 against the reference. \Cref{tab:human} reports Cohen's $\kappa$~\citep{cohen1960kappa}, unweighted for Part~1 and quadratic-weighted for Part~2 (the accuracy judges' mean score rounded to the nearest half point), with 95\% bootstrap intervals over items. Both the response judge and the accuracy judges agree with the annotators more than the annotators agree with one another, and the response judge matches the label given by at least two annotators on \HAOneMajAgree\% of the items. Most disagreements concern responses that give the abstention sentence requested by the prompt and then a premise denial (\HAHybridN{} items); at least two annotators label \HAHybridMajDeny{} of them as premise denials, and without these items the judge's mean $\kappa$ with the annotators is \HAOneJudgeKappaRest. These disagreements affect Correct Abstention Rate, which distinguishes acknowledged uncertainty from premise denial. For comparison, the table also reports agreement after merging abstentions and premise denials into a single category: mean judge--human $\kappa = \HADeclineJudgeKappa$, against \HADeclineHumanKappa{} between annotators. That binary comparison does not validate the distinction required by the headline metric. On accuracy, individual annotators differ mainly in leniency; their mean score (\HATwoHumanMean) is close to that of the accuracy judges (\HATwoPanelMean). Because both parts are sampled evenly across the judges' labels, these values describe agreement on a balanced set rather than on the full distribution of answers.

\begin{table}[!htbp]
\centering
\footnotesize
\setlength{\tabcolsep}{4pt}
\begin{tabular}{@{}lcc@{}}
\toprule
 & Part 1: response & Part 2: accuracy \\
 & $\kappa$ & weighted $\kappa$ \\
\midrule
Human--human & \HAOneHumanKappa{} [\HAOneHumanKappaLo, \HAOneHumanKappaHi] & \HATwoHumanKappa{} [\HATwoHumanKappaLo, \HATwoHumanKappaHi] \\
Judge--human & \HAOneJudgeKappa{} [\HAOneJudgeKappaLo, \HAOneJudgeKappaHi] & \HATwoJudgeKappa{} [\HATwoJudgeKappaLo, \HATwoJudgeKappaHi] \\
Judge--majority & \HAOneMajKappa{} [\HAOneMajKappaLo, \HAOneMajKappaHi] & -- \\
\addlinespace
\multicolumn{3}{@{}l}{\textit{Part 1 with abstentions and premise denials merged}} \\
Human--human & \HADeclineHumanKappa{} [\HADeclineHumanKappaLo, \HADeclineHumanKappaHi] & \\
Judge--human & \HADeclineJudgeKappa{} [\HADeclineJudgeKappaLo, \HADeclineJudgeKappaHi] & \\
\bottomrule
\end{tabular}
\caption{Agreement on the shared annotation set (\HAOneN{} response labels, \HATwoN{} accuracy scores, \HAAnnotators{} annotators): Cohen's $\kappa$ with 95\% bootstrap intervals over items. \emph{Human--human}: mean over annotator pairs. \emph{Judge--human}: mean over annotators of each one's agreement with the response judge (Part~1) or the mean of the accuracy judges (Part~2). \emph{Judge--majority}: against the label at least two annotators gave.}
\label{tab:human}
\end{table}

\DCappendixsection{Prompts}
\label{sec:appendix-prompts}

All prompts are reproduced verbatim from the code that produced the benchmark and the results. Fields in braces, e.g.\ \texttt{\{context\}}, are filled at run time.

\newcommand{\promptfile}[1]{\VerbatimInput[fontsize=\scriptsize, breaklines=true, breakanywhere=true, frame=single, framesep=1.5mm]{#1}}

\subsection{Question generation}
\label{sec:appendix-prompts-generation}
\paragraph{Scenario ($G_\text{scenario}$).}
\promptfile{prompts/gen_scenario.txt}
\paragraph{Fact extraction ($G_\text{facts}$).} \textit{NEW} is the target chunk; \textit{OLD} is the scenario (answerable) or the three nearest $C_\text{pre}$ chunks (unanswerable).
\promptfile{prompts/gen_facts.txt}
\paragraph{Question ($G_\text{question}$).} The same prompt is used for both halves.
\promptfile{prompts/gen_question.txt}
\paragraph{Verification: \textsc{Stated}.}
\promptfile{prompts/verify_fact.txt}
\paragraph{Verification: \textsc{Answers}.}
\promptfile{prompts/verify_answers.txt}

\subsection{Answer generation}
\label{sec:appendix-ans-gen}
\paragraph{With retrieval (single-shot and iterative RAG).} \texttt{\{context\}} lists each chunk with its page title and URL.
\promptfile{prompts/answer_rag.txt}
\paragraph{Without retrieval.}
\promptfile{prompts/answer_norag.txt}
\paragraph{Reflection (iterative RAG).}
\promptfile{prompts/reflection.txt}

\subsection{Labeling and scoring}
\label{sec:appendix-prompts-judges}
\paragraph{Response judge (abstain / deny / answer).}
\promptfile{prompts/judge_outcome.txt}
\paragraph{Accuracy judge.}
\promptfile{prompts/judge_accuracy.txt}
\paragraph{Reference answers (oracle model), answerable questions.}
\promptfile{prompts/oracle_answerable.txt}
\paragraph{Reference answers (oracle model), unanswerable questions (leakage audit).}
\promptfile{prompts/oracle_unanswerable.txt}

\DCappendixsection{Models}
\label{sec:appendix-models}

\Cref{tab:models} lists the evaluated models with their release dates and knowledge cut-offs; \Cref{tab:components} lists the models used to build and score the benchmark, three of which are also evaluated.

\begin{table*}[t]
\centering
\scriptsize
\setlength{\tabcolsep}{3.5pt}
\begin{tabular}{@{}llccll@{}}
\toprule
Family & Model & Switchable & Precision & Released & Knowledge \\
& & thinking & & & cut-off \\
\midrule
Qwen3 \citep{qwen3technicalreport} & 4B, 8B, 14B, 32B, 30B-A3B & yes & 8-bit & Apr 2025$^\dagger$ & not published \\
 & 235B-A22B & yes & 4-bit & Apr 2025$^\dagger$ & not published \\
Qwen3-Next \citep{alibaba2025qwen3next} & 80B-A3B (instruct) & no & 8-bit & Sep 2025$^\dagger$ & not published \\
Qwen3.5 \citep{qwen2026qwen35} & 27B, 35B-A3B, 122B-A10B & yes & 8-bit & Feb 2026$^\dagger$ & not published \\
Gemma 3 \citep{gemma_2025,gemma3card} & 4B, 12B, 27B & no & 8-bit & Mar 2025 & Aug 2024 \\
Gemma 4 \citep{gemmateam2026gemma4} & 31B, 26B-A4B & yes & 8-bit & Apr 2026 & Jan 2025 \\
Llama 3.3 \citep{meta2024llama33} & 70B (instruct) & no & 8-bit & Dec 2024 & Dec 2023 \\
OLMo 3 \citep{olmo2026olmo3} & 32B Think & no & 8-bit & Nov 2025 & Dec 2024 \\
GLM-4.5 \citep{glm2025glm45} & Air & yes & 8-bit & Jul 2025$^\dagger$ & not published \\
Command A \citep{cohere2025commanda,cohere2025commandadocs} & 111B & no & 8-bit & Mar 2025 & Jun 2024 \\
\midrule
\multicolumn{6}{@{}l}{\textit{Models run via OpenRouter}} \\
GPT-4.1 \citep{openai2026modelpages} & mini, nano & no & API & Apr 2025 & Jun 2024 \\
GPT-5 \citep{openai2026modelpages} & mini, nano & no & API & Aug 2025 & May 2024 \\
GPT-OSS \citep{openai2025gptoss} & 120B & no & API & Aug 2025 & Jun 2024 \\
Gemini 2.5 \citep{google2026geminithinking} & Flash & yes & API & Jun 2025 & Jan 2025 \\
 & Flash-Lite & no & API & Jul 2025 & Jan 2025 \\
Gemini 3 \citep{google2025gemini3flash,google2026gemini3guide} & Flash (preview) & no & API & Dec 2025 & Jan 2025 \\
\bottomrule
\end{tabular}
\caption{The 19 open-weight models served locally and the 8 models run through OpenRouter (7 closed, and \texttt{GPT-OSS 120B}). \emph{Switchable thinking}: \emph{yes} if the model was run with thinking both enabled and disabled; models marked \emph{no} were run in one setting, with thinking always on (\texttt{OLMo 3 32B Think}, \texttt{GPT-5}, \texttt{GPT-OSS 120B}), at its low thinking level (\texttt{Gemini 3 Flash}) or without thinking. \emph{Precision}: the weights used in every analysis, 8-bit (FP8) or, for \texttt{Qwen3 235B-A22B}, 4-bit (INT4), the highest available; \emph{API} for models run through OpenRouter. Parameter counts not given by the name: \texttt{GLM-4.5-Air} 106B total and 12B active, \texttt{Gemma 4 26B-A4B} 25.2B and 3.8B, \texttt{GPT-OSS 120B} 117B and 5.1B. Release dates and knowledge cut-offs are taken from the developers' reports, model cards, announcements or API documentation (for closed models, the date of the snapshot called); ``not published'' means none states a knowledge cut-off. $^\dagger$Released after the 31 March 2025 corpus cut-off with no published knowledge cut-off: these models could have seen post-cut-off pages, which the leakage audit (\Cref{tab:leakage}) tests. Gemma 4 and Gemini 3 Flash were also released later, but their published knowledge cut-offs (January 2025) precede our corpus cut-off. The served name and pinned revision of every build are in \texttt{artifacts/v13/served\_models.csv}.}
\label{tab:models}
\end{table*}

\begin{table}[!htbp]
\centering
\footnotesize
\begin{tabular}{@{}>{\raggedright\arraybackslash}p{0.41\columnwidth}>{\raggedright\arraybackslash}p{0.53\columnwidth}@{}}
\toprule
Role & Model \\
\midrule
Embedding (dense retrieval) & Qwen3-Embedding-8B, Q8\_0 \citep{zhang2025qwen3embedding} \\
Reranking (ablation; verifier retrieval) & llama-nemotron-rerank-1b-v2 \citep{nvidia2025nemotronrerank} \\
Question generator & Qwen3.5 122B-A10B, 8-bit \\
Question verifier & Gemma 4 31B, 8-bit \\
Response judge (abstain / deny / answer) & Gemma 4 26B-A4B, 8-bit \\
Accuracy and leakage judges & Gemma 4 31B, Qwen3.8 27B \citep{qwen2026qwen38}, Laguna S 2.1 \citep{poolside2026lagunas21} (8-bit) \\
Reference answers (oracle model) & DeepSeek V4.1 Flash \citep{deepseekai2026deepseekv41flash} \\
\bottomrule
\end{tabular}
\caption{Models used to build and score the benchmark. Three of them are also evaluated (\texttt{Qwen3.5 122B-A10B}, \texttt{Gemma 4 31B} and \texttt{Gemma 4 26B-A4B}); results mark them where they matter.}
\label{tab:components}
\end{table}

\DCappendixsection{Additional Related Work}
\label{sec:appendix-related-work}

\paragraph{RAG vulnerabilities.}
Off-the-shelf RAG pipelines are often suboptimal in specialized domains \citep{barnett2024sevenfailurepointsengineering,tian2025domainspecificdatagenerationframework}, and their accuracy drops when queries vary in formality, readability, politeness or grammar \citep{cao2026styleragsfragilitylinguistic}.
These systems also struggle with knowledge conflicts arising from outdated facts or contradictory evidence \citep{xu2024knowledgeconflictsllmssurvey}.
\citet{longpre-etal-2021-entity} study conflicts between memorized facts and supplied passages; \citet{xie2024adaptive} examine how convincing external evidence and confirmation bias affect which source a model follows.
These studies concern conflicting evidence, whereas \name focuses on missing evidence about documented additions.
Recent studies examine failures to recognize missing evidence \citep{chen2026relevance,setiawan2026promptbased} and questions whose answers are absent from the retrieval corpus \citep{dorosario2026whyrags}.

\paragraph{Hallucination detection and clinical RAG.}
RAGTruth annotates unsupported and contradictory content in generated responses at the word level \citep{niu-etal-2024-ragtruth}.
Almanac evaluates clinical RAG and uses a retrieval-score threshold to trigger abstention when retrieved evidence does not meet that threshold \citep{zakka2024almanac}.
Our evaluation asks whether models acknowledge missing knowledge even when related passages are retrieved.

\paragraph{Self-knowledge.}
Whether a model knows what it does not know has been studied as its ability to recognize unanswerable or unknowable questions \citep{yin-etal-2023-large} and to predict whether it can answer \citep{kadavath2022language}; retrieval can sharpen this awareness of knowledge boundaries \citep{ren-etal-2025-investigating}.
Semantic entropy estimates uncertainty across the meanings of sampled answers \citep{kuhn2023semantic}.
Our no-retrieval condition probes whether models recognize knowledge gaps without external context.
We measure response behavior, not internal confidence or calibration.

\paragraph{Temporal datasets.}
\citet{dhingra-etal-2022-time} introduce TempLAMA and train time-aware language models to represent changing facts and improve calibration on future facts.
Questions about changing facts are usually drawn from news \citep{kasai2024realtimeqawhatsanswer,pmlr-v162-liska22a} or from successive versions of Wikipedia and Wikidata \citep{jang-etal-2022-temporalwiki,tan2023benchmarkingimprovingtemporalreasoning,lin-etal-2025-dynaquest}.
FreshQA adds fast-changing and false-premise questions \citep{vu-etal-2024-freshllms}.
AbstentionBench labels FreshQA questions as requiring abstention when their answers changed after the models' knowledge cut-offs \citep{kirichenko2025abstentionbench}.
ElectionQA23 likewise uses election facts that postdate the knowledge cut-off to evaluate abstention \citep{feng-etal-2024-dont}.
AntiLeak-Bench uses questions about knowledge newer than a model's training corpus to avoid contamination \citep{wu-etal-2025-antileakbench}.
DRAGOn updates a RAG knowledge base over time but does not evaluate abstention \citep{chernogorskii-etal-2026-dragon}.
We use a game's documented update history to obtain discrete, dated additions from a single source.
We audit possible exposure because the dates models actually know can differ from their reported knowledge cut-offs \citep{cheng2024dateddata}.

\paragraph{Minecraft for AI evaluation.}
\textit{Minecraft} is a common testbed for embodied and planning agents \citep{wang2023voyager,fan2022minedojo}.
Plancraft includes intentionally unsolvable crafting tasks and, like MineDojo, offers the \textit{Minecraft} Wiki as a knowledge source \citep{dagan2024plancraftevaluationdatasetplanning,fan2022minedojo}.
We use the wiki's revision history as a dated corpus for retrieval.

\end{document}